\pdfoutput=1
\documentclass[manuscript,authordraft,review=false]{acmart}

\AtBeginDocument{%
  \providecommand\BibTeX{{%
    \normalfont B\kern-0.5em{\scshape i\kern-0.25em b}\kern-0.8em\TeX}}}

\usepackage{amsmath}
\usepackage{amsfonts}       % blackboard math symbols
\usepackage{amsthm}
\usepackage{arydshln}
\usepackage{bold-extra}
\usepackage{cases}
\usepackage{float}
\usepackage[utf8]{inputenc}
\usepackage{multirow}
\usepackage{nicefrac}
\usepackage{subcaption}
\usepackage{tablefootnote}
\usepackage{xargs}                      % Use more than one optional parameter in a new commands

\title{Fairness Through Counterfactual Utilities}

\author{Jack Blandin}
\email{blandin1@uic.edu}
\author{Ian Kash}
\email{iankash@uic.edu}
\affiliation{%
  \institution{University of Illinois at Chicago}
  \city{Chicago}
  \state{IL}
  \country{US}
}

\theoremstyle{definition}
\newtheorem{definition}{Definition}[section]

\theoremstyle{plain}

\theoremstyle{plain}

\theoremstyle{plain}

\theoremstyle{definition}

\newtheorem{assumption}{Assumption}
\theoremstyle{definition}

\counterwithin{table}{section}
\counterwithin{equation}{section}


\begin{document}
    \begin{abstract}
    Group fairness definitions such as Demographic Parity and Equal Opportunity make assumptions about the underlying decision-problem that restrict them to classification problems.
    Prior work has translated these definitions to other machine learning environments, such as unsupervised learning and reinforcement learning, by implementing their closest mathematical equivalent.
    As a result, there are numerous bespoke interpretations of these definitions.
    Instead, we provide a generalized set of group fairness definitions that unambiguously extend to all machine learning environments while still retaining their original fairness notions.
    We derive two fairness principles that enable such a generalized framework.
    First, our framework measures outcomes in terms of \textit{utilities}, rather than predictions, and does so for both the decision-algorithm and the individual.
    Second, our framework considers \textit{counterfactual} outcomes, rather than just observed outcomes, thus preventing loopholes where fairness criteria are satisfied through self-fulfilling prophecies.
    We provide concrete examples of how our \textit{counterfactual utility} fairness framework resolves known fairness issues in classification, clustering, and reinforcement learning problems.
    We also show that many of the bespoke interpretations of Demographic Parity and Equal Opportunity fit nicely as special cases of our framework.
\end{abstract}
    \maketitle
    \section{Introduction}\label{section:intro}

Machine learning (ML) is used to automate decision-making in settings such as hospital resource allocation \cite{obermeyer2019dissecting}, job application screening \cite{raghavan2020mitigating}, and criminal sentencing recommendations \cite{kleinberg2018human}.
Given the high social impact of these settings, the field of \textit{fairness in machine learning} has gained significant attention in recent years.
In this work, we focus on \textit{group fairness} definitions, where an algorithm is considered fair if its results are independent of one or more protected attributes such as gender, ethnicity, or sexual-orientation.
Many group fairness works focus only on classification settings~\cite{berk2018fairness, chouldechova2017fair, corbett2017algorithmic, dwork2012fairness, NIPS2016_9d268236, NIPS2017_a486cd07, galhotra2017fairness}.
This often conceals assumptions that do not always hold true in other contexts, resulting in definitions that are tightly coupled with a particular problem domain.
In this paper we examine four such assumptions.

\begin{assumption}\label{assumption:fair_predictions_fair_outcomes}
    Fair predictions have fair outcomes.
\end{assumption}
Many group fairness definitions require equal predictions between protected groups~\cite{berk2018fairness, chouldechova2017fair, corbett2017algorithmic, dwork2012fairness, NIPS2016_9d268236}.
%and generally consider the binary case with a \textit{minority} group and a \textit{majority} group.
For example,
in the binary case with a \textit{minority} group and a \textit{majority} group,
\textit{Demographic Parity} considers a binary classifier to be fair if it predicts the positive class for individuals in the minority group and majority groups with equal probability.
This implicitly assumes that a positive prediction is always a good outcome for an individual.
However, there are many problem domains where this is not true.
For instance, \cite{liu2018delayed} consider an algorithm that predicts whether or not a loan applicant will repay a loan, which then informs a loan-approval decision.
In this scenario, a positive prediction results in a loan approval, which has a positive outcome for those who will pay back the loan, but has a \textit{negative} outcome for those who will default on the loan.
More generally, in situations where predictions impact individuals from the minority and majority groups differently, prediction-based fairness definitions may actually result in unfair outcomes.
We refer to this as the \textit{prediction-outcome disconnect} issue.

\begin{assumption}\label{assumption:decision_influence_outcome}
    Observed values of the target variable are independent of predictions.
\end{assumption}
Some fairness definitions depend on the observed value of the target variable as well as the prediction.
For example, Equal Opportunity
%conditions on qualified individuals, 
requires equal treatment of the qualified individuals in each group,
where \textit{qualified} refers to individuals who were observed to be in the positive class \cite{NIPS2016_9d268236}.
However, consider a classifier that predicts if an individual convicted of a crime will recidivate, where the prediction informs a judge's decision on whether to impose a prison sentence.
It is possible that the decision of whether to assign prison time actually influences the individual's probability of being \textit{qualified}, which corresponds to not recidivating.
For example, suppose there is a group of \textit{backlash} individuals 
%within the minority group 
that will only recidivate if they are sentenced to prison \cite{imai2020principal}.
If the algorithm predicts that these individuals will recidivate, which causes the judge to sentence them to prison, these individuals will be considered \textit{unqualified} because they will in fact recidivate.
However, if the algorithm had instead predicted these backlash individuals to not recidivate, then they will not actually recidivate and will be considered qualified.
Thus an algorithm can satisfy Equal Opportunity through a \textit{self-fulfilling prophecy} by manipulating who is considered qualified.

\begin{assumption}\label{assumption:objective_predict_target}
    The objective is to predict some unobserved target variable.
\end{assumption}
In classification
%fairness 
problems, the goal is to make a single prediction of some latent qualification attribute of the individual.
% This assumes that there exists some latent qualification attribute and that there is only a single prediction made per individual.
% Thus, it is reasonable to define \textit{qualified} individuals as those observed to be in the positive class.
However, this is not true in other ML environments where the decision is not necessarily a prediction of some ground-truth value, and where there may be more than one decision per individual.
In sequential decision settings such as reinforcement learning (RL), the goal is to maximize a reward rather than predict a target.
Additionally, there can be multiple sequential decisions made for each individual and we may wish to measure fairness across the entire sequence.
Although some attempts have been made to translate group fairness to the sequential decision setting~\cite{bower2017fair,wen2021algorithms}, they assume a specific problem structure which limits their application.
Similarly, there are also several bespoke translations in clustering problems ~\cite{chen2019proportionally,abbasi2021fair,NEURIPS2019_fc192b0c,NIPS2017_978fce5b}, each of which is tied to a variant of what fairness means in their particular context.
% With so many bespoke definitions to choose from, it makes it difficult to standardize group fairness across problem domains.

\begin{assumption}\label{assumption:single_prediction}
    Decisions for one individual do not impact other individuals.
\end{assumption}
Each classification prediction is independent of the predictions made for other individuals.
However, this does not generalize to all of ML.
In clustering, for instance, the impact of one individual's cluster assignment may depend on the cluster assignments of other individuals.
For example, \citeauthor{abbasi2021fair}~\cite{abbasi2021fair} consider redistricting as a fair clustering problem, where fairness implies that constituents from each political party are equally represented by their assigned district.
In order to measure how well a constituent is represented by their district, we need to know \textit{who else} was assigned to their district.
We term this \textit{conjoined fairness} when the impact of a decision for one individual requires measuring the decisions made for other individuals as well.
Conjoined fairness can also arise in other settings such as ranking (if the impact of being ranked second depends on who was ranked first) or RL (if, e.g., the decision to hire an individual may preclude the future hiring of another).

\subsection{Our Contributions}
In this work, we provide a more general group fairness framework that does not rely on the aforementioned assumptions, thereby allowing it to extend to a wide variety of classification, clustering, and RL tasks.
We show that our definitions encompass the standard classification definitions as well as several of their domain-specific adaptations.
% We demonstrate a repeated pattern of (1) defining the utilities of the individual and decision-maker, (2) establishing the fairness objective, and (3) showing how the fairness objective can be defined in exclusively terms of our framework's parameters.
We also demonstrate how our definitions can capture domain idiosyncrasies
% as variables and parameters
through appropriate instantiation which reduces the need for bespoke definitions.

We focus on Demographic Parity and Equal Opportunity, but as we discuss in Appendix \ref{appendix:otherdefs}, our framework can extend to other group fairness definitions as well.
There are two principles that differentiate our framework: \textit{welfare} and \textit{counterfactual utility outcomes}.

\paragraph{Welfare}
In order to resolve the prediction-outcome disconnect issue, it is instructive to consider the intuition behind Demographic Parity, which requires that the probability that an individual receives a beneficial outcome is independent of the individual's protected attribute.
We can resolve the prediction-outcome disconnect issue by measuring the individual's outcome \textit{directly}.
Thus, we introduce a new variable called \textit{welfare}, which represents the individual's utility resulting from a prediction.
For example, a loan applicant that is likely to pay back a loan has positive welfare if the algorithm predicts that the applicant will pay back the loan, since the prediction will likely result in a positive outcome for the applicant.
On the other hand, a loan applicant that will likely default on a loan has negative welfare if the algorithm predicts that they will pay back the loan, since this prediction will likely result in a negative outcome (defaulting).
Our Demographic Parity and Equal Opportunity definitions require a \textit{welfare function} to be defined for the given problem domain, and then measure equal welfare instead of equal predictions.
By measuring fairness directly in terms of welfare, our definitions enforce fair outcomes even in domains where the predictions impact individuals differently.
Furthermore, since \textit{utility} is a more generally applicable concept than \textit{prediction} or \textit{target variable}, this approach makes sense in a broader range of domains where Assumptions~\ref{assumption:objective_predict_target} and~\ref{assumption:single_prediction} may not hold.

\paragraph{Counterfactual utilities}
We saw in our discussion of Assumption~\ref{assumption:decision_influence_outcome} that the standard definition of Equal Opportunity is vulnerable to self-fulfilling prophecies.  
In order to remedy this, we construct a more general Equal Opportunity definition by giving a more general interpretation of what it means to be \textit{qualified}.
As we explain in Section \ref{subsection:counterfactual_qualification}, we interpret qualification as an individual where there exists a decision that will yield a good outcome for both the decision-algorithm \textit{and the individual}.
In other words, we measure qualification in terms of counterfactual utility outcomes for both the decision-algorithm and the individual.
By considering counterfactual outcomes, our Equal Opportunity definition prevents self-fulfilling prophecies and is well-defined for a broader range of ML environments.

\subsection{Related Work}

%Prior work on Utility
Previous work has incorporated notions of utility or welfare into fairness problems in machine learning~\cite{jabbari2017fairness,liu2018delayed,kim2020preference} and such approaches are common in economics~\cite{finocchiaro2021bridging}.  However, this work has not formulated group fairness definitions in terms of utilities.
As the lone exception, \cite{wen2021algorithms} recently and independently introduced the idea of using welfare to generalize group fairness definitions to Markov decision processes (MDPs).  However, they do not consider the possibility of generalizing to other domains such as clustering or make use of counterfactual utilities, so their approach applies only to a restricted class of MDPs.
See Section~\ref{subsection:applications_objective_not_predict_target} for more discussion.

%Prior work on Counterfactual
Our use of counterfactuals may seem reminiscent of the literature on causal fairness notions such as counterfactual fairness~\cite{NIPS2017_a486cd07, kilbertus2017avoiding, nabi2018fair,loftus2018causal,makhlouf2020survey}.
However, there the counterfactual is what decision the algorithm would make if the protected attribute were different, while for us the counterfactual is what a different choice of algorithm would do.
% %Our approach is compatible with these definitions.
\citeauthor{krishnaswamy2021fair}~\cite{krishnaswamy2021fair} consider counterfactual algorithm choices, but do so to have a baseline on how well the best classifier for a group can perform.
Our use of counterfactuals is more similar to the way they are used in principal fairness \cite{imai2020principal}; 
%where potential outcomes (counterfactual) are considered rather than observed (factual) outcomes.
see Section~\ref{subsection:applications_pred_infl_outcomes} and Appendix \ref{appendix:worked_example_recidivism_prediction} for more discussion.
Our notion of optimizing for counterfactual outcomes with the prediction algorithm itself as input is similar to performative prediction~\cite{perdomo2020performative, miller2021outside} which studies how to find an optimal and stable prediction algorithm when predictions influence the observed outcome.

Previous work has explored relaxing each of our four assumptions, although typically in isolation.  This includes work on prediction-outcome disconnect ~\cite{liu2018delayed,creager2020causal}, self-fulfilling prophecies ~\cite{imai2020principal}, fairness in sequential decision-making for
reinforcement learning~\cite{jabbari2017fairness} and pipelines~\cite{bower2017fair,dwork2020individual,emelianov2019price}, 
fair ranking~\cite{celis2017ranking,singh2019policy,zehlike2021fairness},
and fair clustering~\cite{NIPS2017_978fce5b,NEURIPS2019_fc192b0c,chen2019proportionally,abbasi2021fair}.
    \section{Preliminaries}\label{section:preliminaries}

The group fairness
%notions 
definitions
we study were originally developed in the context of classification.
%problems.  
Following \citeauthor{NIPS2016_9d268236}~\cite{NIPS2016_9d268236}, we think of this task as predicting a target value $Y$ based on features $X$ and protected attribute $Z$ where the population of individuals is represented by the joint distribution of $(X,Z,Y)$ and the goal is to develop a classifier $\hat{Y}(X,Z)$.
We typically omit the arguments to $\hat{Y}$ for brevity when they are clear.
%A particular individual
An individual
%(formally outcome of the random variable $(X,Z,Y)$)
is an element of $\mathcal{X} \times \mathcal{Z} \times \mathcal{Y}$.
Here $\mathcal{X}$ and $\mathcal{Y}$ are the sets of possible feature values and target values.
We restrict the protected attribute space to be binary $\mathcal{Z}=\{0,1\}$ purely for ease of exposition.\footnote{
    Appendix \ref{appendix:more_than_two_protected groups} considers situations with $|\mathcal{Z}|>2\;$.}
We refer to individuals with $Z{=}0$ as the \textit{minority} group, and those with $Z{=}1$ as the \textit{majority} group.
There is a loss function $L:\mathcal{Y}\times\mathcal{Y}\rightarrow\mathbb{R}$ and the objective is to find the classifier that minimizes expected loss $L(Y,\hat{Y}(X,Z))$.
We refer to the tuple $(X,Z,Y,L)$ as a supervised learning classification problem (SLCP).  

While there are many group fairness definitions~\cite{pessach2020algorithmic,verma2018fairness}, we focus our exposition on two of the most important to illustrate our approach.
\footnote{Appendix \ref{appendix:otherdefs} details how other fairness metrics are implemented in our framework.}

\begin{definition}[Classification Demographic Parity]
    A classifier $\hat{Y}$ satisfies Classification Demographic Parity (\texttt{DemParClf}) for an SLCP $(X,Z,Y,L)$ if 
        \begin{equation}%\small
            P(\hat{Y}{=}1\mid Z{=}0) = P(\hat{Y}{=}1\mid Z{=}1) \\ \label{eq:trad_dempar}
        \end{equation}
    \label{def:dem_parity}
\end{definition}

\begin{definition}[Classification Equal Opportunity]
    A classifier $\hat{Y}$ satisfies Classification Equal Opportunity (\texttt{EqOppClf}) for SLCP $(X,Z,Y,L)$ if
        \begin{equation}%\small
            P(\hat{Y}{=}1 \mid Y{=}1, Z{=}0) = P(\hat{Y}{=}1 \mid Y{=}1, Z{=}1) \label{eq:trad_eqopp} \; .
        \end{equation}
    \label{def:eq_opp}
\end{definition}
    \section{Intuition for Counterfactual Utility}\label{section:intuition}
In this section we provide an intuitive explanation of our approach for generalizing group fairness definitions beyond classification.
We defer a formal treatment to Section~\ref{section:definitions}.

%\subsection{Fairness over Welfare}\label{subsection:welfare_function}
\subsection{Fairness Through Welfare}\label{subsection:welfare_function}

\texttt{DemParClf} is defined exclusively in terms of SLCP variables.
However, the concept behind Demographic Parity, that equal outcomes should be enforced across groups, may be relevant in any domain.
Suppose that we instead define a more general version of Demographic Parity where we replace $\hat{Y}$ with a variable $W$ that represents the \textit{welfare} of the decision from the individual's perspective.
Assuming $W$ can take on a range of values, our general Demographic Parity becomes
\begin{equation}%\small
    P(W\geq\tau \mid Z{=}0) = P(W\geq\tau \mid Z{=}1), \label{eq:informal_dem_par}
\end{equation}
where $\tau \in \mathbb{R}$ is some domain-specific threshold representing the minimum welfare to be considered a \textit{good} outcome for the individual.
Rather than assuming that a prediction of 1 is a good outcome, as in \texttt{DemParClf}, we use $W$ to explicitly capture the relationship between a decision and an outcome, which allows us to incorporate a variety of domain-specific aspects.
Additionally, rather than require equal expected $W$, we elect to enforce equal probabilities that $W$ exceeds a threshold $\tau$.\footnote{
    Non-threshold design choices are reviewed in Appendix~\ref{appendix:threshold_alternatives}. }
We do so for two reasons.
First, it provides the closest translation of \texttt{DemParClf}, which aligns with our objective of expanding existing fairness definitions to other environments.
Second, it decouples decisions on "how individuals are impacted" from "what is considered an acceptable outcome", with the former defined by $W$, and the latter defined by $\tau$.
These two concepts may be orthogonal, so decoupling them induces more focused discussions when designing fairness problems, as well as allows for $W$ to be shared across domains with similar impact dynamics but differing thresholds for acceptable outcomes.
%For example, the effects of an employee's salary on their purchase power may be similar across two locations, but thresholds for what is considered an acceptable amount of purchase power may differ.
For example, it might be natural to use a person's income as a measure of their welfare, but thresholds for what is considered an acceptable income may vary across locations.
% Under this formulation of Demographic Parity, an algorithm is fair if the probability that a good welfare outcome is achieved is equal for both values of the protected attribute.
% Requiring equal welfare instead of equal predictions enables this definition to be relevant in any setting where $W$ and $Z$ can be defined.

\subsection{Counterfactual Utility Qualification}\label{subsection:counterfactual_qualification}
We can also modify \texttt{EqOppClf} to use $W$ instead of $\hat{Y}$: $P(W\geq\tau \mid Y{=}1,Z{=}0)=P(W\geq\tau \mid Y{=}1,Z{=}1)\;$.
However, this definition is still using the SLCP variable $Y$.
In order to extend this definition to environments outside of classification, we need to inspect the intuition for Equal Opportunity, which is that the probability that a \textit{qualified} individual receives a beneficial outcome is independent of the individual's protected attribute.
The part of the definition referring to the \textit{beneficial outcome} is already covered by the welfare concept, so we only need to modify the definition to allow \textit{qualified} to also to extend to other settings.
We develop intuition for what it means to be qualified by considering some examples of Equal Opportunity:
    \begin{itemize}
        \item The probability that a \textit{skilled} job candidate is hired is independent of their protected attribute.
        \item The probability that a \textit{straight-A} student is admitted to a university is independent of their protected attribute.
    \end{itemize}
Thus, qualified individuals are those whose beneficial outcome also benefits the decision-algorithm:
    \begin{itemize}
        \item The beneficial outcome for a job applicant is to be hired. If hired, a skilled job candidate will also benefit the employer since they will be competent at their job.
        \item The beneficial outcome for a student is to be admitted to the university. If admitted, a straight-A student will benefit the university by enhancing the university's reputation.
    \end{itemize}
Therefore, our general Equal Opportunity interpretation is:
\begin{quote}
     For the subset of individuals \textit{where there exists an outcome that will benefit both the individual and the decision-algorithm}, the probability that a beneficial individual outcome occurring is independent of the individual's protected attribute.
    %  \footnote{
    %     Although our general interpretation of Equal Opportunity conditions on individuals that can produce beneficial outcomes for both the individual and the decision-algorithm, it only requires equal probability of beneficial \textit{individual} outcomes.
    %     However, we can easily construct alternative definitions that require beneficial outcomes for the lender as well using our framework.
    %     Therefore, our more general Equal Opportunity does not guarantee that a good decision-algorithm occurs.
    %     Rather, the set of decision-algorithms that satisfy our Demographic Parity but do not satisfy our Equal Opportunity, are bad according to the decision-algorithm.
    %  }
\end{quote}
We can represent this in equation form as
\begin{equation}%\small
    P(W\geq\tau\mid\Gamma{=}1,Z{=}0)=P(W\geq\tau\mid\Gamma{=}1,Z{=}1) \label{eq:informal_eq_opp}
\end{equation}
where $\Gamma$ is an indicator random variable with $\Gamma = 1$ when the decision-algorithm can produce an outcome that is beneficial for both the individual and the decision-maker.
The benefit to the individual is captured by $W$.  We can similarly capture the impact on the decision-maker with a \textit{cost function} $C$, thus $\Gamma$ becomes:\footnote{
    We use \textit{cost} over \textit{utility} due to the convention of minimizing loss functions.}
\begin{equation}%\small
    \Gamma=\begin{cases}
        1 \; \text{ if } \; \exists \hat{Y}' : W_{\hat{Y}'} \geq \tau \land C_{\hat{Y}'} \leq \rho \\
        0 \; \text{ otherwise} \; .
    \end{cases}
    \label{eq:informal_eqopp_gamma}
\end{equation}
Here $W_{\hat{Y}'}$ and $C_{\hat{Y}'}$ are the expected welfare and cost, respectively, produced by predictor $\hat{Y}'$ and $\rho$ is similar to $\tau$ but for the cost.
We can validate that Equation \ref{eq:informal_eq_opp} generalizes well by applying it to the Section \ref{section:intro} recidivism example where \texttt{EqOppClf} allows for self-fulfilling prophecies:
\begin{quote}
     For the subset of individuals (inmates) where there exists an outcome that will benefit both the individual (no prison) and the decision-algorithm (no recidivism), the probability that a beneficial individual outcome (no prison) occurring is independent of the individual's protected attribute.
\end{quote}
We see that our more general Equal Opportunity resolves the self-fulfilling prophecy issue by conditioning on individuals who \textit{could have been qualified}.
Thus, the qualified individuals are those that will not recidivate if they do not receive prison time.
In other words, our more general interpretation conditions on \textit{counterfactually} qualified individuals.
% There are many other such counterfactuals where it is natural to condition on in this manner; see Section \ref{subsection:applications_pred_infl_outcomes} for an example and Appendix~\ref{appendix:otherdefs} for additional discussion.
    \section{Counterfactual Utility Fairness}\label{section:definitions}

We now provide our formal model for defining welfare and counterfactual qualification.
We do so in an abstraction that we term a \textit{Fairness Decision-Making Problem} (FDMP).
FDMPs generalize the classification definitions from Section~\ref{section:preliminaries} to other ML environments such as RL and clustering, and enables these environments to resolve the issues outlined in Section \ref{section:intro}.
% deal with domain aspects such as prediction-outcome disconnect, self-fulfilling prophecies, and conjoined fairness.

In a Fairness Decision-Making Problem (FDMP), a decision-maker selects a decision-algorithm $m$ which has somehow been selected from a class of such algorithms $M$.
An individual is an outcome of random variable $(I,Z)$.
$\;I\in\mathcal{I}$ represents the individual's non-sensitive attributes that are relevant for determining a decision's impact on the decision-maker or on the individual themselves.
E.g. in university admissions, $I$ may include the applicant's GPA since it is may be a proxy for post-graduation success which impacts the university's reputation; $I$ may also include the applicant's family income level since a rejection may have greater impact for applicants with less options to choose from.
$Z\in\mathcal{Z}=\{0,1\}$ captures the individual's protected attribute.
%We assume that the protected attribute is binary $\mathcal{Z}=\{0,1\}$.
The decision-maker has a \textit{cost function} $\;C:(\mathcal{I}\times \{0,1\})\times M\rightarrow\mathbb{R}$ which maps an individual's relevant non-sensitive attributes, sensitive attribute, and a decision-algorithm to the expected cost.
We capture the cost associated with a given decision-algorithm $m$ as a random variable $C_m:\mathcal{I}\times\{0,1\}\rightarrow\mathbb{R}$.
The impact of $m$ on an individual is captured by the \textit{welfare function} $W:(\mathcal{I}\times\{0,1\})\times M\rightarrow\mathbb{R}$ which is identical to the cost function except that it maps to expected welfare instead.
Similar to the cost function, $W$ depends on the individual's attributes and the decision-algorithm, so we represent the welfare associated with a given decision-algorithm $m$ as a random variable $W_m:\mathcal{I}\times\{0,1\}\rightarrow\mathbb{R}$.
Two threshold constants $\tau$ and $\rho$ are required where $\tau$ represents the minimum welfare needed for the outcome to be considered \textit{good} from the individual's perspective, and $\rho$ represents the maximum cost needed for the outcome to be considered \textit{good} from the decision-algorithm's perspective.%
%\footnote{
    %As we show in Section \ref{section:applications}, in practice there is often a clear distinction between what is a good outcome and what is a bad outcome, so the choices for $\tau$ and $\rho$ are obvious.
    %In cases where the distinction is less clear, we can compute $\tau$ by taking the median $W$ produced by a fair-naive algorithm that only optimizes for efficiency.
    %$\rho$ can be computed the same way except with $C$ instead of $W$.
%    }
\footnote{
    In Section \ref{section:applications} we consider a number of examples which show how these thresholds can be chosen based based on domain-specific considerations.
}
In summary, an FDMP is compactly represented by a 7-tuple $(I, Z, M, W, C, \tau, \rho)$.
Fairness definitions are then characterized by comparisons of welfare $W$, cost $C$, thresholds $\tau$ and $\rho$, and protected attribute $Z$.

We can now formally define our Welfare Demographic Parity and Counterfactual Utility Equal Opportunity.
\begin{definition}[Welfare Demographic Parity]
    Given FDMP $(I, Z, M, W, C, \tau, \rho)$, a decision-algorithm $m\in M$ satisfies Welfare Demographic Parity (\texttt{DemParWelf}) if
    \begin{equation}%\small
        P (W_m\geq \tau \mid Z{=}0 ) = P (W_m\geq \tau \mid Z{=}1 )
        \label{eq:fdmp_dempar}
    \end{equation}
\end{definition}

\begin{definition}[Counterfactual Utility Equal Opportunity]
Given FDMP $(I, Z, M, W, C, \tau, \rho)$, a decision-algorithm $m\in M$ satisfies Counterfactual Utility Equal Opportunity (\texttt{EqOppCfUtil}) if
    \begin{equation}%\small
        P (W_m\geq \tau \mid \Gamma{=}1, Z{=}0 ) = P (W_m\geq \tau \mid \Gamma{=}1, Z{=}1 )
        \label{eq:fdmp_eqopp}
    \end{equation}
    where $\Gamma$ is an indicator variable with
    \begin{equation}%\small
        \Gamma=\begin{cases}
            1 \; \text{ if } \; \exists m' \in M : W_{m'} \geq \tau \land C_{m'} \leq \rho \\
            0 \; \text{ otherwise}
        \end{cases}
        \label{eq:fdmp_eqopp_gamma}
    \end{equation}
\end{definition}

\subsection{Counterfactual Utility Applied to Binary Classification}
Now we show how to apply our generalized definitions to a specific environment: binary classification.
A binary classification problem is an SLCP $(X,Z,Y,L)$.
%with $L(Y,\hat{Y})=Y(1-\hat{Y}) + \hat{Y}(1-Y)$.\footnote{
%    We use zero-one loss, but other loss functions also work.}
For concreteness we assume $L$ is the zero-one loss.
We can construct a corresponding FDMP with
    non-sensitive individual attributes $I=(X\times Y)$,
    decision algorithms space $M = \hat{\mathcal{Y}}=(X\times \{0,1\})\times\{0,1\}$ the set of all possible classifiers,
    and cost function $C=L$.
    %We assume that 
    A positive prediction $\hat{Y}=1$ always implies a good outcome for the individual.
    This corresponds to a welfare function $W=\hat{Y}$ with minimum threshold $\tau=1$.
    We set the maximum cost threshold to $\rho=0$ so that a good outcome from the decision-maker's perspective reflects a correct prediction ($L=0$).
    %We also assume that
    The individual's target $Y$ is not influenced by the prediction $\hat{Y}$, thus the parameterized welfare $W_m=W=\hat{Y}$ and $C_m=C=L$.
A binary classifier $\hat{Y}:\mathcal{X}\times\{0,1\}\rightarrow\{0,1\}$ under this problem formulation satisfies \texttt{DemParWelf} if
{ %\small
\begin{align}
    P(\hat{Y}\geq 1 \mid Z{=}0)=P(\hat{Y}\geq1 \mid Z{=}1) \label{eq:app_sl_dempar}
\end{align}}%
Because this is a binary classification problem, $\hat{Y}\geq1$ is equivalent to $\hat{Y}=1$, which makes Equation \ref{eq:app_sl_dempar} equivalent to \texttt{DemParClf} (Equation \ref{eq:trad_dempar}).
Similarly, a classifier $\hat{Y}$ satisfies \texttt{EqOppCfUtil} if
{%\small
\begin{align}
    P(\hat{Y}\geq1 \mid \Gamma{=}1,Z{=}0)=P(\hat{Y}\geq1 \mid \Gamma{=}1,Z{=}1) \label{eq:app_sl_eqopp}
\end{align}}%
where $\Gamma$ is an indicator variable with
\begin{align}%\small
    \Gamma=\begin{cases}
        1 \; \text{ if } \; \exists \hat{Y}' \in \hat{\mathcal{Y}}:
            \hat{Y}'\geq 1 \land %\operatorname{zeroOneLoss}
            L(Y,\hat{Y}') \leq 0 \\
        0 \; \text{ otherwise}
    \end{cases} \label{eq:app_sl_eqopp_gamma}
\end{align}
%where $\hat{\mathcal{Y}}=(X\times \{0,1\})\times\{0,1\}$ is the set of all possible classifiers.
If $\Gamma{=}1$ in Equation \ref{eq:app_sl_eqopp_gamma}, then it must be that $Y{=}1$, which makes Equation \ref{eq:app_sl_eqopp} equivalent to traditional Equal Opportunity (Equation \ref{eq:trad_eqopp}).
Since Equation \ref{eq:app_sl_dempar} reduces to Equation \ref{eq:trad_dempar}, and Equation \ref{eq:app_sl_eqopp} reduces to \ref{eq:trad_eqopp}, \texttt{DemParClf} and \texttt{EqOppClf} are special cases of \texttt{DemParWelf} and \texttt{EqOppCfUtil}, respectively.

    \section{Counterfactual Utility in Practice}\label{section:applications}
% * 1st par: we've given intuition for these solutiosn but haven't discussed how they're a practical solution.
In Section \ref{section:intro} and \ref{section:intuition} we discussed the motivation and intuition for our counterfactual utility fairness framework.
In Section \ref{section:definitions} we defined the framework and showed that classification Demographic Parity and Equal Opportunity are special cases of our framework.
However, apart from the classification definitions being special cases, we have yet to demonstrate that our framework is useful in practice.
% This leads us to ponder what properties are desirable in a fairness framework.
% 2nd par: (currently 1st paragraph). Okay so why is this a practical solution. One thing you'd want is that it exists. We have some arguments about why this isn't trivial, and why it's a good thing. Okay so it's not going to be an existence theorem.
For instance, a natural question to ask is whether there exist decision-algorithms that satisfy Definitions~\ref{eq:fdmp_dempar} and~\ref{eq:fdmp_eqopp}.  It is easy to construct trivial examples where they do not: simply choose a policy space $M$ that does not include any fair decision-algorithms.  More interestingly, even if we choose a rich $M$ that includes all reasonable decision-algorithms, fair ones can fail to exist.  Consider a $W$ such that $W_m = W_{m'}$ for all $m,m' \in M$.  That is, the decision-algorithm has no effect on welfare.  Then unless the scenario is already ``naturally'' fair (i.e. it is inherent in $W$ that $P (W_m\geq \tau \mid Z{=}0 ) = P (W_m\geq \tau \mid Z{=}1)$) no fair decision-algorithm exists.  We view this as a virtue of our definition; if the class of decision-algorithms considered by $M$ is too ``weak'' in its effects to be able to generate fairness then our definition is correct to say that none of the decision-algorithms suffice to do so.  In such a case it might be better to consider interventions not in $M$. Alternatively, we can use our definitions to quantify the degree of unfairness of different options. (See Section~\ref{subsection:applications_preds_dont_imply_eq_outcomes} for an example.)

% * 3rd par: So what are we doing instead in this section. (curently 3rd paragaph)
% Despite the dependence on the details of $M$ and $W$ for whether fair decision-algorithms exist, in many natural cases they do.
While the existence of fair policies may depend on the technical details of $M$ and $W$, in many natural cases fair policies do exist.
We examine such scenarios in Sections \ref{subsection:applications_preds_dont_imply_eq_outcomes}, \ref{subsection:applications_pred_infl_outcomes}, \ref{subsection:applications_objective_not_predict_target}, and \ref{subsection:applications_decision_does_not_impact_others}.
In each of these scenarios, one or more of the four assumptions detailed in Section \ref{section:intro} are violated, and we demonstrate the necessity for our framework in order to adequately measure fairness.
% In these,  we point out specific examples where our definitions identify the ``correct'' $m$ as fair, and where existing definitions do not.
We also point out a number of prior definitions which are special cases of our framework, some of which have existence results~\cite{NIPS2017_978fce5b} and real-world applications~\cite{wen2021algorithms, NEURIPS2019_fc192b0c, NIPS2017_978fce5b}, showing that it is possible to establish our definitions in cases of practical interest.

\subsection{Experiment: Prediction-Outcome Disconnect with German Credit Dataset}\label{subsection:applications_preds_dont_imply_eq_outcomes}

Here we provide an experimental analysis on an environment where classification fairness metrics fail to appropriately measure fairness due to Assumption \ref{assumption:fair_predictions_fair_outcomes}.
We then demonstrate how our utility-based fairness definitions from Section \ref{section:definitions} resolve the issue.
In order to ensure that our analysis is consistent with other group fairness works, we leverage the \textit{fairness-comparison} benchmark from \cite{friedler2019comparative} for data preprocessing, algorithm implementation, and fairness measurement calculations.\footnote{
    We make some modifications to the framework, but these are only for extension purposes, such as for new fairness measurements. 
    % Add this once accepted.
    The full code repository is available at https://github.com/jackblandin/fairness-comparison.
    } 

\paragraph{Dataset}
We consider the loan application scenario described by the German Credit Dataset \cite{Dua:2019}, which consists of 1,000 loan application records.
Each record in the dataset consists of 20 attributes about a loan applicant, including a binary label indicating whether the applicant is a good or bad credit risk.
Following convention~\cite{friedler2019comparative}, we consider the credit risk label as our prediction target $Y$, with $Y=1$ corresponding to good-risk applicants and $Y=0$ corresponding to bad-risk applicants.
The classification objective is to correctly predict which applicants are good-risk, and which are bad-risk.
In addition to the credit risk label, the dataset consists of other financial attributes about the applicant such as the number of open credit lines, credit history, as well as demographic information such as age and sex.
For this experiment, we consider the applicant's sex the protected attribute, with $Z=0$ corresponding to female applicants and $Z=1$ corresponding to male applicants.
The dataset also provides a payoff matrix representing the downstream "cost" of each prediction error, where we assume that loans are granted to applicants predicted to be good-risk, and loans are rejected for applicants predicted to be bad-risk.
We can use this payoff matrix to define our FDMP \textit{cost function} $C$:
    \begin{equation}
        C(Y, Z, \hat{Y}) = \begin{cases}
            	0 \;\;\; \text{if} \; \hat{Y}=Y \\
            	1 \;\;\; \text{if} \; \hat{Y}=0 \land Y=1 \\
            	5 \;\;\; \text{if} \; \hat{Y}=1 \land Y=0 \; \\
            \end{cases}
        \label{eq:german_cost_fn}
    \end{equation}
where $\hat{Y}$ is the binary prediction with $\hat{Y}=1$ if the classifier predicts a good-risk applicant and $\hat{Y}=0$ if it predicts bad-risk.
% Following convention~\cite{friedler2019comparative}, we consider consider good-risk applicants to be those who will repay a loan, and bad-risk as those who will default.
% Therefore, for some applicant $(X_i,Y_i)$, we assume that a $\hat{Y}(X_i)=1$ prediction results in granting a loan to the applicant, and that $\hat{Y}=0$ results in rejecting them.
Equation \ref{eq:german_cost_fn} implies zero cost for granting loans to good-risk applicants or rejecting bad risk applicants, a cost of 1 if a good-risk applicant is rejected for a loan, and a cost of 5 if a loan is granted to a bad-risk applicant.
The dataset thus articulates that it is much worse for an applicant to fail to repay a loan (i.e. defaulting) than it is to reject an applicant that would have repaid a loan.

\paragraph{Algorithms}
We evaluate four of the binary classification algorithms studied in \cite{friedler2019comparative}, with a mix of algorithms that optimize for accuracy only and algorithms that optimize for both accuracy and fairness.%\footnote{
%    By \textit{efficiency}, we refer to the primary utility being optimized; e.g. the loss function in classification, or the reward function in RL.
%    In our FDMP framework, efficiency is represented by the cost function.}
The first two algorithms are standard classification techniques that only optimize for accuracy: Decision Tree (\texttt{DT}) and Support Vector Machine (\texttt{SVM}).
The remaining two algorithms, Feldman Decision Tree (\texttt{Feld-DT})~\cite{feldman2015certifying} and Feldman SVM (\texttt{Feld-SVM})~\cite{feldman2015certifying}, optimize for both accuracy and fairness by preprocessing techniques that modify the input attributes $X$ to have equal marginal distributions based on the subsets of that attribute with a given sensitive value.
% \texttt{ZafarFair} adds a fairness constraint to the efficiency optimization.

\paragraph{Fairness measures}
We evaluate each of the four algorithms on three different fairness measures.
The first two are classification-based measures and the third is a utility based measure that leverages our definitions from Section \ref{section:definitions}.
The first fairness measure is \texttt{DemParClfRatio}, which represents the extent to which \texttt{DemParClf} is achieved:
    ~\footnote{
        \texttt{DemParClfRatio} is typically referred to as "Disparate Impact". We use "DemParClfRatio" to make it clear that it is a quantification of \texttt{DemParClf}.
    }
    % ~\footnote{
    % In our figures we show $1-|DI|$ instead of $DI$ so that a higher numerical value implies a higher "fairness" value, with perfect fairness achieved when $1-|DI| = 0$.
    % We follow this convention for the other two fairness measures as well.
    % }
    \begin{equation}
        \texttt{DemParClfRatio} = \min\biggl(
            \frac{P(\hat{Y}=1 \mid Z=0)}{P(\hat{Y}=1 \mid Z=1)},
            \frac{P(\hat{Y}=1 \mid Z=1)}{P(\hat{Y}=1 \mid Z=0)},
        \biggr)
        \label{eq:disparate_impact} \; .
    \end{equation}
We compute the minimum of the minority-majority ratio and its inverse so that a higher value implies a higher level of fairness, with perfect fairness achieved at $\texttt{DemParClfRatio}=1$.
We follow this convention for the remaining fairness measures.
The second fairness metric represents the extent to which \texttt{EqOppClf} is achieved:
    \begin{equation}
        \texttt{EqOppClfRatio} = \min\biggl(
            \frac{P(\hat{Y}=1 \mid Y=1, Z=0)}{P(\hat{Y}=1 \mid Y=1, Z=1)},
            \frac{P(\hat{Y}=1 \mid Y=1, Z=1)}{P(\hat{Y}=1 \mid Y=1, Z=0)}
        \biggr)
        \label{eq:eqopp_ratio} \; .
    \end{equation}
% In this particular dataset, false positives (applicant defaults) have the most significant impact on individuals.
% Because \texttt{EqOppClf} does not consider false positives in its calculation, we also include a third classification metric, called \texttt{EqOppFPRatio}, that is equivalent to \texttt{EqOppClfRatio} except that it measures false positives (FPs) instead of true positives (TPs):
%     \begin{equation}
%         \texttt{EqOppFPRatio} = \frac{P(\hat{Y}=1 \mid Y=0, Z=0)}{P(\hat{Y}=1 \mid Y=0, Z=1)}
%         \label{eq:eqopp_fp_ratio} \; .
%     \end{equation}
% \texttt{BCRRatio} as a baseline comparison to our utility-based fairness measures.
% The third fairness measure is also a classification measure, and is defined as the ratio of the minority and majority balanced classification rates (BCR), which we refer to as \texttt{BCRRatio}:
% % \texttt{BCRRatio} is the ratio of the sums of the true positive rates (TPR) and true negative rates (TNR) of the minority and majority:
%     \begin{equation}
%         \texttt{BCRRatio} = \frac{P(\hat{Y}=1 \mid Y=1, Z=0)+P(\hat{Y}=0 \mid Y=0, Z=0)}{P(\hat{Y}=1 \mid Y=1, Z=1)+P(\hat{Y}=0 \mid Y=0, Z=1)}
%         \label{eq:z_bcr_ratio} \; .
%     \end{equation}
% \texttt{BCRRatio} considers a classifier as fair if it has equal (balanced) accuracy for each protected group, which makes it a good complement to \texttt{DI} that only compares prediction rates.
% \texttt{BCRRatio} is similar in concept to \texttt{EqOppClf}, but is actually more encompassing since \texttt{EqOppClf} only considers positive samples.
For the third measure, we introduce a utility-based measure, called \texttt{DemParWelfRatio}, which follows our threshold-based utility definitions from Section \ref{section:definitions}:
    \begin{equation}
        \texttt{DemParWelfRatio} = \min\biggl(
            \frac{P(W \geq \tau \mid Z=0)}{P(W \geq \tau \mid Z=1)},
            \frac{P(W \geq \tau \mid Z=1)}{P(W \geq \tau \mid Z=0)}
        \biggr)
        \label{eq:threshwelfdi}
    \end{equation}
where the welfare function $W$ is equal to the (negative) cost function defined in Equation \ref{eq:german_cost_fn},\footnote{
    Since the applicant and lender (prediction algorithm) share the same incentives, we can reasonably assume that they share the same utility functions.
    Section \ref{subsection:applications_pred_infl_outcomes} discusses an example where the welfare and cost functions diverge.
    % A future line of work might consider evaluating how these fairness measures change as the welfare and cost functions diverge.
} and $\tau=-1$ so that $W\geq \tau$ indicates the applicant did not default.
We selected this threshold since it best separates the extreme welfare values that occur from applicant defaults.
\texttt{DemParWelfRatio} can be interpreted as the extent to which \texttt{DemParWelf} is achieved.

\begin{figure*}
    \centering
    \frame{\includegraphics[scale=.35]{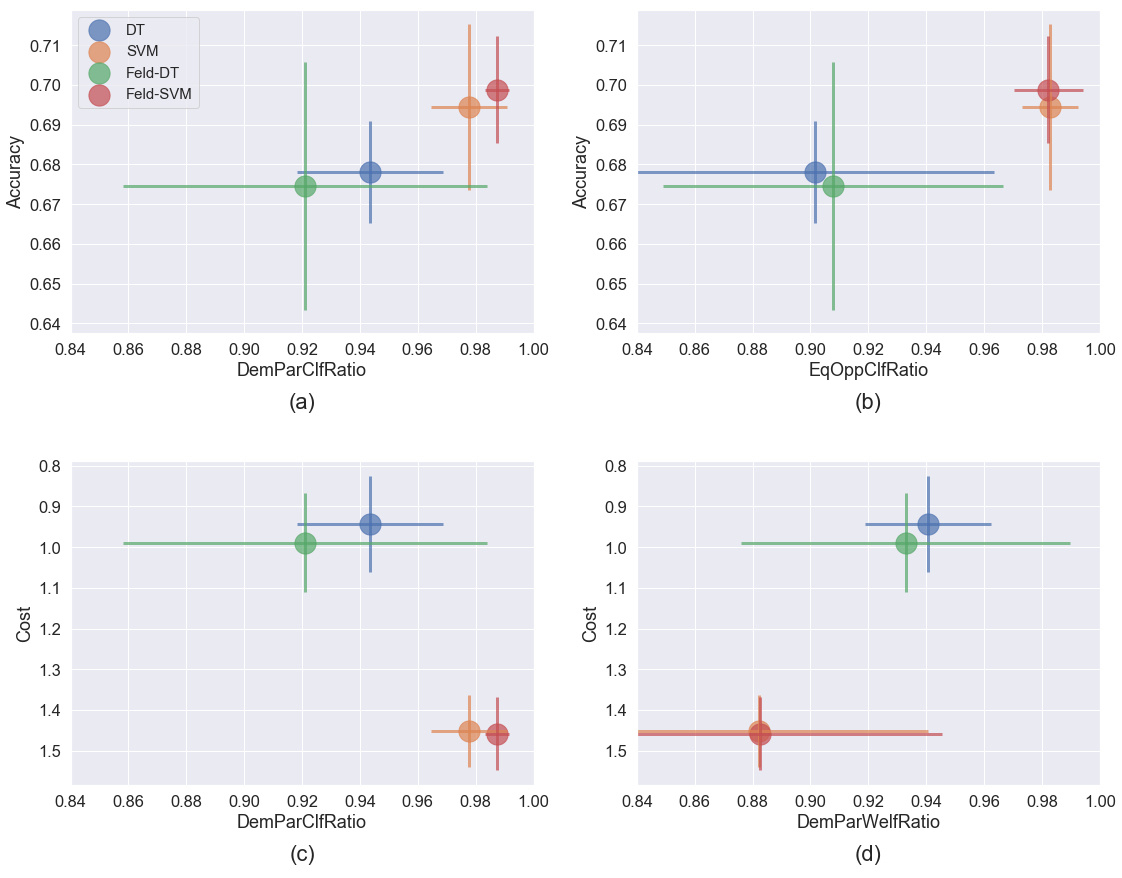}}
    \caption{
        Using the German Credit dataset, four binary classification algorithms are evaluated for efficiency (y-axis) and fairness (x-axis).
        The filled circles represent the average measurements over 10 iterations of k-folds, and the horizontal and vertical lines represent the 10th and 90th percentiles.
        Plots (a) and (b) compare accuracy and classification fairness measures; (c) compares compares cost, which is equivalent to negative welfare in this case, against a classification fairness measure, and (d) compares cost against our utility-based fairness measure.
        The axes are constructed so that more accurate (efficient for (c) and (d)) algorithms are higher-up and more fair algorithms are further right.
    }
    \label{fig:germancredit_3plots}
\end{figure*}

\paragraph{Results}
%In order to measure both the expected value and variance of each performance measure, each of the five algorithms are trained over 10 iterations of k-folds and their measures recorded for each fold.
We execute and measure each algorithm using 10-fold cross-validation.
For each performance measurement, we report the average value as well as the 10th and 90th percentiles.
The results are shown in Figure \ref{fig:germancredit_3plots}.
Figures \ref{fig:germancredit_3plots}a and \ref{fig:germancredit_3plots}b indicate that the SVM algorithms are more accurate and more fair than the Decision Tree algorithms according to both classification-based fairness measures.
Therefore, the lender would not need to make a decision about a fairness-efficiency trade-off since the SVM algorithms are better with respect to both.
However, a reasonable lender would likely prefer to use cost rather than accuracy for its efficiency measure.
Figure \ref{fig:germancredit_3plots}c shows that the Decision Trees algorithms are more cost-efficient, which suggests that choosing one of these four algorithms requires making a trade-off between fairness and efficiency.
But if the lender cares more about cost than accuracy, and the lender and applicants have similar incentives, then it also makes sense to measure fairness in terms of cost, as in \texttt{DemParWelfRatio}.\footnote{
        Technically, \texttt{DemParWelfRatio} is measured in terms of welfare, not cost. But cost is equal to negative welfare in this case, so \texttt{DemParWelfRatio} is indeed a cost-based measure.
        }
Figure \ref{fig:germancredit_3plots}d shows the Decision Tree algorithms outperform the SVM algorithms on both cost and cost-based fairness.
%so the lender can avoid a trade-off.
This means that despite being less accurate they produce fewer defaults and more evenly distribute those defaults between male and female applicants.  
Thus, in contrast to classification-based definitions, our approach reveals that a lender that correctly realizes that raw accuracy is not the most important metric in this setting does not actually face a trade-off between fairness and efficiency when selecting among these four models.

This experiment demonstrates the harm that can arise from fairness metrics that make Assumption \ref{assumption:fair_predictions_fair_outcomes}, where prediction equality is favored over impact equality.
Without properly considering the downstream impact of predictions, which can be done by defining our FDMP parameters, even a well-intentioned decision-maker can inadvertently implement unfair algorithms or face a false trade-off between fairness and efficiency.

\subsection{Example: Self-Fulfilling Prophecies}\label{subsection:applications_pred_infl_outcomes}
Here we illustrate how Assumption \ref{assumption:decision_influence_outcome} allows for self-fulfilling prophecies with \texttt{EqOppClf} but not with \texttt{EqOppCfUtil}.
We use the recidivism prediction example posed by Imai and Jiang~\cite{imai2020principal} where a binary classifier predicts whether an inmate convicted of a crime will recidivate, which informs a judge's decision $\hat{Y}$ of whether to detain ($\hat{Y}=0$) or release ($\hat{Y}=1$) the inmate.
The target variable $Y$ corresponds to whether or not the inmate will recidivate, with $Y=0$ indicating recidivism.
This problem differs from typical classification since $Y$ is influenced by $\hat{Y}$.
When decisions influence the observed target variable, it is helpful to visualize the dataset by \textit{principal strata} ~\cite{frangakis2002principal} where each principal stratum characterizes how an individual would be affected by the decision $\hat{Y}$ with respect to the variable of interest $Y$. %~\cite{imai2020principal}.
Since this is a binary classification problem with binary decisions and binary targets, we have a total of four principal strata.
We assign labels to each stratum according to their behavior in Table \ref{tab:recidivism_stratum}.
For example, an individual in the \texttt{Backlash} stratum will recidivate if they are detained, so $P(Y{=}1 \mid \hat{Y}{=}0)=0$, but will not recidivate if released, so $P(Y{=}1 \mid \hat{Y}{=}1)=1$.%
%For simplicity, we assume the principal stratum of each individual is already known.
% \footnote{
    % In practice, the principal stratum of an individual is not known, but can be estimated using techniques from the causal inference literature \cite{rubin2005causal}.
    % Additionally, we can also use off-policy evaluation techniques \cite{bang2005doubly, creager2020causal} in order to evaluate counterfactual scenarios without explicitly learning the principal strata.
    % }

%Since an inmate will always prefer to be released over being detained, we consider the welfare function to be
To model inmates who always prefer to be released, we can take the welfare function to be
a binary function with $W=1$ when the inmate is released and $W=0$ when detained.
Similarly, to model a judge (decision-maker) who always prefers outcomes where the inmate does not recidivate, we can set the cost function to $C=1$ when the inmate recidivates and $C=0$ when they do not.
Following a similar fairness criteria of that posed by \cite{imai2020principal}, we want to ensure that inmates who will not recidivate if released are released with equal probability for each protected group.
Therefore, we set the welfare threshold $\tau=1$ so that a good outcome from an inmate's perspective is when they are released.
Similarly, we set the cost threshold $\rho=0$ such that a good outcome from the judge's perspective is when an inmate does not recidivate.

\texttt{EqOppClf} considers an individual as qualified if the value of $Y$ is observed to be $1$.
This means that an inmate is qualified if  they do not recidivate, which corresponds to inmates that (a) are in the \texttt{Safe} stratum, (b) are in the \texttt{Backlash} stratum and are released, or (c) are in the \texttt{Preventable} stratum and are detained.
Therefore, even if the minority and majority inmate populations are identical in every way other than their protected attribute, a classifier could satisfy \texttt{EqOppClf} while having different release rates for inmates who would not recidivate.
For example, a decision-maker could get away with detaining more safe minority inmates than the majority simply by releasing more preventable inmates.
We can see how this works by inspecting an equivalent form of \texttt{EqOppClf}:
    \begin{equation*}%\small
        \frac{\texttt{Count}(\hat{Y}=1 \cap Y=1 \cap Z=0)}{\texttt{Count}(Y=1 \cap Z=0)}
        = \frac{\texttt{Count}(\hat{Y}=1 \cap Y=1 \cap Z=1)}{\texttt{Count}(Y=1 \cap Z=1)} \; . \\
    \end{equation*}
Detaining more \texttt{Safe} minority inmates reduces the numerator for the minority, but this can be offset by releasing more \texttt{Preventable} inmates which causes the minority denominator to also decrease.
The classifier thus attains "fairness" through a self-fulfilling prophecy by manipulating who is considered "qualified".

Alternatively, \texttt{EqOppCfUtil} does not allow for self-fulfilling prophecies since it has a prediction-independent definition of "qualification".
Referencing Equation \ref{eq:fdmp_eqopp_gamma}, an individual is qualified under \texttt{EqOppCfUtil} with $\tau=1$ and $\rho=0$ if $\exists \hat{Y'} \in \hat{\mathcal{Y}} : W_{\hat{Y}'} \geq 1 \land C_{\hat{Y}'} \leq 0\;$.
In other words, an inmate is considered qualified if there exists a classifier that will produce $W\geq 1$ and $C\leq 0$, which is only possible for individuals who will not recidivate when released.
Thus, according to \texttt{EqOppCfUtil}, an inmate is qualified if they are in the \texttt{Safe} or \texttt{Backlash} stratum, regardless of if they are detained or released.
For a complete worked example with population numbers for each of the principal strata, see Appendix \ref{appendix:worked_example_recidivism_prediction}.

% Other fairness definitions can also prevent self-fulfilling prophecies, such as Imai and Jiang's \textit{Principal Fairness} \cite{imai2020principal}, which requires equal predictions across each principal stratum.
% As we show in Appendix \ref{appendix:worked_example_recidivism_prediction}, our FDMP framework easily accommodates Principal Fairness.
% \texttt{EqOppCfUtil} is not the only fairness definition that can prevent self-fulfilling prophecies.
% For instance, Imai and Jiang propose \textit{Principal Fairness}~\cite{imai2020principal}, which requires equal predictions across each principal stratum.
% Although Principal Fairness can be implemented in our framework, as we show in \ref{appendix:worked_example_recidivism_prediction}, we prefer \texttt{EqOppCfUtil} since Principal Fairness can be overly strict.
% E.g. Principal Fairness requires equal release rates for Dangerous, Backlash, Preventable, and Safe inmate.
% the release rate to be equal across Backlash groups and the Safe release rates to be equal across groups, rather than requiring the release rates of the aggregate Backlash and Safe to be equal.

\begin{table*}
    \small
    \def\arraystretch{1.2}%  1 is the default, change whatever you need
    \begin{center}
    \begin{tabular}{c|c|c|c} 
        & & $P(Y=1|\hat{Y}=1)=0$ & $P(Y=1|\hat{Y}=1)=1$ \\ \hline
        & & Dangerous & Backlash \\
        % & & \\[-1em] % Adds space between a single row
        \multirow{2}{*}{$P(Y=1|\hat{Y}=0)=0$} & Detained & $\texttt{Unq}$ & $\texttt{CfUtil}$ \\
        & Released & $\texttt{Unq}$ & $\texttt{Clf}, \texttt{CfUtil}$ \\ \hdashline
        & & Preventable & Safe \\
        % \\[-1em] % Adds space between a single row
        \multirow{2}{*}{$P(Y=1|\hat{Y}=0)=1$} & Detained & $\texttt{Clf}$ & $\texttt{Clf}, \texttt{CfUtil}$ \\
        & Released & $\texttt{Unq}$ & $\texttt{Clf}, \texttt{CfUtil}$ \\
    \end{tabular}
    \end{center}
    \caption{
        The four principal strata for the recidivism prediction problem.
        Cells with \texttt{Clf} correspond to qualified inmates according to \texttt{EqOppClf}.
        \texttt{CfUtil} are qualified according to \texttt{EqOppCfUtil}.
        \texttt{Unq} are unqualified according to both.
    }
    \label{tab:recidivism_stratum}
\end{table*}

\subsection{Example: Fairness in Reinforcement Learning}\label{subsection:applications_objective_not_predict_target}

In this section we provide an example of how our counterfactual utility definitions extend to environments beyond classification.
Specifically, we apply them to RL, a domain that violates Assumption \ref{assumption:objective_predict_target}.
We provide formalisms for MDPs and then discuss how to construct the corresponding FDMP.
We provide a worked RL example in Appendix \ref{appendix:worked_example_two_stage_loan_mdp}.

\begin{definition}[Markov decision process]\label{def:mdp}
A Markov Decision Process (MDP) is a 6-tuple $\{S,\mathcal{A},T,R,\gamma, \mu\}$ where
    $S$ is a set of states;
    $A$ is a set of actions;
    $T : S \times \mathcal{A}\rightarrow \Delta S$ is a mapping of state-action pairs to a distribution over new states:  $T(s_t|s_{t-1},a_{t-1})\;$;
    $R: S \times \mathcal{A}\rightarrow \mathbb{R}$ is the reward function, which maps a state-action pair to a real number;
    $\gamma \in [0,1]$ is the discount factor;
    and $\mu$ is the initial state probability distribution.
    A typical goal is to find a policy $\pi^* \in \Pi$ that maximizes the expected discounted reward.
\end{definition}

When constructing the FDMP, some parameters can be inferred from the MDP directly, while others need to be defined according to the problem domain and desired fairness criteria.
FDMP parameters $I$, $Z$, $M$, and $C$ can be inferred from the MDP as follows.
The MDP state $s\in S$ corresponds to an individual's unprotected attributes $\tilde{s}\in \tilde{S}$ (i.e. $I= \tilde{s}$) and protected attribute $Z$ (i.e. $s=\{\tilde{s},Z\}$ and $S=\tilde{S}\times \mathcal{Z}$).
Therefore, the initial state $s_0$ represents an individual in the first timestep $s_0=\{\tilde{s}_0,Z\}$ .
The individual's unprotected attributes $\tilde{s}$ can change over subsequent timesteps, and do so according to the transition function $s_{t}\sim T(s_{t-1},\pi(s_{t-1}))$ with $s_{t}=\{\tilde{s}_{t},Z\}$.
We assume that the individual's protected attribute $Z$ does not change throughout an episode.
$M = \Pi$, the set of policies.
Following RL reward convention, the cost function $C_\pi$ is the negative expected cumulative sum of rewards after executing the policy $\pi\in\Pi$:
\begin{equation}%\small
    C_\pi = - 
    %\underset{s_t\sim T(s_{t-1}, \pi(s_{t-1}))}
    %{\mathbb{E}} \bigl[R(s_0,\pi(s_0)) + \sum_{t=1}^{\infty} \gamma^t R(s_t,\pi(s_t))\bigr]\label{eq:cost_function_rl} \; .
    {\mathbb{E}} \bigl[ \sum_{t=0}^{\infty} \gamma^t R(s_t,\pi(s_t))\bigr]\label{eq:cost_function_rl} \; .    
\end{equation}
We can construct the welfare function $W$ similarly by defining it as the expected cumulative sum of a domain-defined \textit{welfare contribution function} $w:S\times\mathcal{A}\rightarrow\mathbb{R}$:
\begin{equation}%\small
    W_\pi = 
    %\underset{s_t\sim T(s_{t-1}, \pi(s_{t-1}))}
    %{\mathbb{E}}\bigl[w(s_0,\pi(s_0)) + \sum_{t=1}^{\infty}\gamma^t w(s_t,\pi(s_t))\bigr] \label{eq:welfare_function_rl}
    {\mathbb{E}}\bigl[ \sum_{t=0}^{\infty}\gamma^t w(s_t,\pi(s_t))\bigr] \label{eq:welfare_function_rl}
\end{equation}
Thus, in order to define $W(s,\pi)$, and therefore $W_\pi(s)$, we only need to define $w(s,a)$.
As we can see, the welfare contribution function $w$ shares the same signature as the reward function $R$, and can be thought of as the \textit{individual's} reward function.
In essence, our counterfactual utility framework  requires the RL practitioner to construct 
%a second reward function that corresponds to the individual's utility.
two reward functions rather than the standard one.
Since the goal of group fairness definitions is to ensure some level of outcome equality for groups of individuals, then constructing a mapping of policy actions to individual outcomes (i.e. utility) is \textit{necessary} in order to adequately measure fairness.\footnote{
        This statement is true for all ML environments, not just RL.
    }
The remaining FDMP parameters $\tau$ and $\rho$ can then be assigned so as to implement the desired fairness criteria.

We reserve a fully worked example to Appendix \ref{appendix:worked_example_two_stage_loan_mdp} where we apply Equations \ref{eq:cost_function_rl} and \ref{eq:welfare_function_rl} to Equations
\ref{eq:fdmp_dempar},
\ref{eq:fdmp_eqopp}, and \ref{eq:fdmp_eqopp_gamma} in order to define 
%\texttt{EqOppCfUtil} in the RL setting.
%We then compare \texttt{EqOppCfUtil} to that of \cite{wen2021algorithms} which also provides an RL-translation of Equal Opportunity using an individual utility function, but does not fully consider counterfactual scenarios.
\texttt{DemParWelf} and \texttt{EqOppCfUtil} in the RL setting.
We then compare our definitions to those of
\citeauthor{wen2021algorithms}~\cite{wen2021algorithms} who recently independently extended the notions of demographic parity and equal opportunity to the RL setting.  They also use a utility function so their translation of demographic parity to this setting is equivalent to ours (i.e. their approach is a special case of our framework).  However, their RL-translation of Equal Opportunity  does not fully consider counterfactual scenarios and we argue that \texttt{EqOppCfUtil} is superior.

\subsection{Example: Conjoined Fairness in Clustering}\label{subsection:applications_decision_does_not_impact_others}

Here we show how Assumption \ref{assumption:single_prediction} results in conjoined fairness issues in the clustering setting.
% We also demonstrate that bespoke problem domains do not require bespoke fairness definitions; they only require bespoke \textit{welfare} and \textit{cost} functions.
Additionally, we show that that two different clustering fairness definitions, \textit{balanced clustering}~\cite{NIPS2017_978fce5b} and \textit{representative clustering}~\cite{abbasi2021fair}, correspond to \texttt{DemParWelf} with different welfare function implementations.
%First we provide relevant definitions.

A \textit{clustering problem} is a 4-tuple ($\bar{X}, \bar{Z}, \mathcal{K}, L)$ where
    $\bar{X}$ is an $n$-length vector of unprotected individual attributes with
    %$\bar{X}^{0\leq i\leq n-1}$ representing the unprotected attributes of the $i^\text{th}$ individual in $\bar{X}$;
    $\bar{X}^{i}$ representing the unprotected attributes of the $i^\text{th}$ individual
    %in $\bar{X}$ $(0 \leq i \leq n-1)$;
    in $\bar{X}$;    
    $\bar{Z}$ is an $n$-length vector of protected individual attributes;
    $\mathcal{K}$ is the number of clusters;
    % \iffalse{
    % $L:(\mathcal{X}\times\mathcal{Z})\times \mathcal{J}\rightarrow\mathbb{R}$ is the loss function which maps the set of individuals $(\mathcal{X}\times\mathcal{Z})$ and their clustering assignments 
    % %$K$
    % (where $\mathcal{J}=(\mathcal{X}\times\mathcal{Z})\times\mathcal{K}$ is the space of possible \textit{clusterings})
    %     %where $\mathcal{X}$ is the set of possible individual majority attributes ($\bar{X}^i\in \mathcal{X} \; \forall \; 0 \leq i \leq n$),
    %     %$\mathcal{Z}$ is the set of possible individual majority attributes ($\bar{Z}^i\in \mathcal{Z} \; \forall %\; 0\leq i \leq n$),
    %     %and $\mathcal{J}:(\mathcal{X}\times\mathcal{Z})\times\mathcal{K}$ is the space of possible \textit{clusterings} for the set of individuals ($\mathcal{X}\times\mathcal{Z}$).
    % The objective of a clustering problem is to find a clustering $J\in \mathcal{J}$ that minimizes the loss $L(\bar{X},\bar{Z},J)$.
    % }\fi
    A clustering $J$ maps the dataset of individuals $(\bar{X},\bar{Z})$ to an $n$-length vector of cluster assignments; we use $J(X^i,Z^i)$ to denote the cluster to which individual $i$ is assigned.  
    The loss function 
    %L:(\bar{X}\times\bar{Z})\times \mathcal{J}\rightarrow\mathbb{R}$
    $L(\bar{X},\bar{Z},J)$
    maps the set of individuals and their clustering assignments to a real number.
    In clustering, we have a fixed dataset of individuals, so to represent this as a FDMP we have the random variable $(I,Z)$ sample an individual at random from the dataset.
    
%% Explain redistricting example.
As a motivating example, we consider a clustering problem where the goal is to segment a geographic region into $\mathcal{K}$ districts (clusters), where each district is represented by a single elected official.
We assume a two-party system where the individual's protected attribute reflects their political party affiliation.
We wish to evaluate Demographic Parity for a given set of district boundaries $J$.
% An intuitive translation of \texttt{DemParClf} to this setting requires that the probability that a constituent is assigned to a positive district is equal for both parties, where a \textit{positive} district implies that a constituent assigned to this district would consider it a good outcome.
% However, we cannot know which districts are \textit{positive} unless we know who else was assigned to that district.
% The problem is that \texttt{DemParClf} does not capture \textit{conjoined fairness} notions where the fairness of a decision for an individual depends on the decisions made for other individuals.
%Another
One interpretation of Demographic Parity is \textit{balanced clustering}~\cite{NIPS2017_978fce5b} where each political party is required to be evenly
%balanced across
split among
all clusters.
We can implement this using our framework with a welfare function equal to the proportion of the constituent's political party for a given district.
%:
%\begin{equation}
%    W_J(k) = {\operatorname{count}(j^k,\bar{Z}^i)}/{\sum_{z\in\mathcal{Z}}\operatorname{count}(j^k,z)} \label{eq:welfare_balanced_clustering}
%\end{equation}
%where
    %$j^k$ is the set of constituents who are assigned to district $k$ by $J$,
    %and $\operatorname{count}(j^k,Z)$ is the number of constituents in $j^k$ that are in political party $\bar{Z}^i$.
If individual $i$ is sampled and $k = J(X^i,Z^i)$ we have 
\begin{equation}%\small
    W_J = \frac{|\{ j \mid J(X^j,Z^j)=k \wedge Z^i = Z^j \}|} {|\{ j \mid J(X^j,Z^j)=k \}|}
    \label{eq:welfare_balanced_clustering}
\end{equation}
Therefore, a clustering $J$ satisfies \texttt{DemParWelf} if
\begin{equation}%\small
    P(W_J\geq\tau \mid \bar{Z}^i{=}0) = P(W_J\geq\tau \mid \bar{Z}^i{=}1) %\; \text{for } k=\{0,1,...,\mathcal{K}-1\}
    \label{eq:demparwelf_clustering}
\end{equation}
where $\tau$ is the minimum proportion of the constituent's party needed to be considered a good outcome for the individual (e.g. the proportion of that party in the population).
% When $\tau=.5$ this is identical to their notion of a \textit{perfectly} balanced clustering, which requires an equal number of constituents from both political parties in every district. $\tau=0$ corresponds to their \textit{fully unbalanced} clustering notion where each district consists of constituents from a single political party.
% Results in gerrymandering.
Unfortunately, balanced clustering results in the same political party ratio across all districts, which is actually a form of gerrymandering where one political party controls all of the districts by maintaining a slight majority in each.
This can be remedied with \textit{representative clustering}~\cite{abbasi2021fair}.
This definition uses a similarity function that maps a constituent and their district assignment to their similarity value, and requires that constituents from each party have equal similarity.\footnote{
    It is possible to construct a similarity function that produces gerrymandering.
    This is an example of a broader issue where poorly-chosen fairness notions might be worse than nothing.
    % Therefore, it should be defined so that individuals with similar political beliefs have high similarity scores.
}
In our FDMP framework, we can implement a thresholded version of representative clustering by setting the welfare function equal to the similarity function:
\begin{equation}%\small
    W_J = \operatorname{similarity}((\bar{X}^i,\bar{Z}^i),J(\bar{X}^i,\bar{Z}^i)) \label{eq:welfare_representative_clustering}
\end{equation}
where $(\bar{X}^i,\bar{Z}^i)$ is the constituent and $J(\bar{X}^i,\bar{Z}^i))$ is the constituent's district assignment.
% A clustering $J$ satisfies \texttt{DemParWelf} for representative clustering if it satisfies Equation \ref{eq:demparwelf_clustering} with $W_J$ equal to Equation \ref{eq:welfare_representative_clustering}.
Under representative clustering,  $\tau$ represents the minimum constituent-district similarity for the district to be considered a good representation of the constituent.
Therefore, if a clustering $J$ satisfies \texttt{DemParWelf} with $W_J$ equal to Equation \ref{eq:welfare_representative_clustering}, then a constituent from either political party will have equal probability of being well-represented by their assigned district,
%which will prevent gerrymandering.
preventing gerrymandering.
As this example illustrates, rather than providing an entirely new fairness definition for a bespoke fairness problem, we can instead leverage the generality of \texttt{DemParWelf} and 
%restrict domain idiosyncrasies to just $W$ and $\tau$.
capture domain idiosyncrasies 
through the definitions of $W$ and $\tau$ even when fairness is conjoined.
    \section{Discussion}\label{section:conclusion}

Using our FDMP framework, we have proposed generalizations of standard group fairness definitions such as Demographic Parity and Equal Opportunity based on two principles: using utility functions to capture outcomes for both individuals and the decision maker, and considering counterfactual outcomes.
We have shown that our definitions subsume the standard definitions from  classification settings~\cite{NIPS2016_9d268236} as well as several of their domain-specific translations such as in RL and clustering.
%\cite{NIPS2017_978fce5b,abbasi2021fair}
Furthermore, we demonstrated how our definitions reduce the need for bespoke definitions since domain idiosyncrasies are captured as FDMP parameters and variable definitions.

% We conclude by discussing the limitations and potential negative societal impacts of our work.
% While we have shown examples of how our framework can address problems such as prediction-outcome disconnects, self-fulfilling prophecies, and conjoined fairness and we believe these examples are representative of large classes of situations, we do not know how to prove, or even state, a claim that our framework resolves all such issues.  Our framework lets us naturally map group fairness definitions between settings and flexibly capture domain features, but this does not mean that any particular definition is sensible for any particular setting.  A poorly-chosen instantiation of our framework might be worse than not using it at all due to degraded performance, introducing new versions of bias or unfairness, or providing false confidence that fairness issues have been addressed.

We conclude by discussing four practical considerations.
First, there may be situations where counterfactual outcomes need to be considered but the causal structure between decisions and observed outcomes is unknown.
In this case,
%practitioners
we
can leverage techniques from the causal inference literature~\cite{rubin2005causal} to estimate the causal structure, or use off-policy evaluation techniques \cite{bang2005doubly, creager2020causal} in order to estimate counterfactual outcomes without explicitly learning the causal structure.
Second, our analysis focuses on expanding the range of settings where group fairness definitions can be applied. There are additional issues that are orthogonal to our goal, including how best to satisfy fairness during learning and how to trade off between fairness and utility.
Third, practitioners may disagree on the most appropriate welfare function.
Disagreement on the welfare function may also occur using prediction-based definitions, but will manifest as disagreement of whether a particular fairness definition is suitable for a particular problem domain.
This may inadvertently rule out fairness definitions as unsuitable simply because they are not well-defined.
We showed an example of this in Section \ref{subsection:applications_preds_dont_imply_eq_outcomes} where Demographic Parity was unsuitable using the prediction-based definition (\texttt{DemParClf}), but suitable when well-defined with our framework (\texttt{DemParWelf}).
We actually see welfare disagreements as a benefit of our framework since, unlike prediction-based fairness metrics, our definitions naturally decouple discussions of "how individuals are impacted" from "what is fair", thereby focusing debates on the actual points of disagreement.\footnote{
    This is similar to our Section \ref{subsection:welfare_function} discussion of decoupling "how individuals are impacted" and "what is an acceptable individual outcome" with separate variables for $W$ and $\tau$.
    Except here we are referring to decoupling both $W$ and $\tau$ from the fairness definition.
    I.e. our framework decouples "how individuals are impacted", "what is an acceptable individual outcome", and "what is fair".
    }
Finally, an important potential negative societal impact of our work, as with any other fairness definition, is the potential for misuse.
Our framework lets us naturally map group fairness definitions between settings and flexibly capture domain features, but this does not mean that any particular definition is sensible for any particular setting.
A poorly chosen instantiation may be worse than not using it at all due to degraded performance, newly introduced bias or unfairness, false confidence that fairness issues have been addressed, or over-optimization of equality at the cost of efficiency.

% Practicality
%     * Practicality of learning the causal structure of the data in order to infer counterfactual outcomes
%         * Footnote in 5.2 mentions that we can leverage techniques from causal inference literature. We should probably bring this up again.
%     * Practicality of gathering personalized welfare functions in order to construct welfare function.
%         * We should make a note immediately after when we define the welfare and cost functions in (Section 4) that clarifies that although an individual is an input to the functions, we generally don't need to define it differently for each individual in practice. Footnote? 
%     * Disagreement between how to define the utility functions.
%         * Good item to include in the Discussion section. Mention that this is the same issue as defining the Reward function in RL or the loss function in supervised learning.
%     * Thresholds needlessly dichotomize individuals’ welfare.
%         * We already give a pointer to Appendix
%         * Address this in Review document
        
% How should you use group fairness definitions broadly
%     * A  poorly-chosen  instantiation  of  our  framework  might  beworse than not using it at all due to degraded performance,introducing new versions of bias or unfairness, or providingfalse confidence that fairness issues have been addressed.
%     * Solutions may not be /efficient/. (issue with fairness in general)
%     * We do not address how to combine fairness and other tradeoffs (equality vs. efficiency).
%         * Include citations to other work that addrss those things.
    % \clearpage
    \bibliographystyle{ACM-Reference-Format}
    % \bibliography{sample-base}
    \bibliography{references}
    \clearpage
    \appendix
    \section{Appendix}\label{section:appendix}
% \section{Extension to Other Group Fairness Definitions.}

\subsection{Extension to Other Group Fairness Definitions.}
\label{appendix:otherdefs}
In addition to Demographic Parity and Equal Opportunity, we can extend our framework to implement various other group fairness definitions.
In this section, we provide a partial list of these implementations. 
Furthermore, we observe that each of the counterfactual utility implementations of these group fairness definitions can be constructed as expressions of $W_m$, $\tau$, $Z$, and $\Gamma$ alone.

\paragraph{Equalized Odds}
Similar to Equal Opportunity, \textit{Equalized Odds} \cite{NIPS2016_9d268236} requires both the true positive rates ($P(\hat{Y}=1|Y=1)$) and false positive rates ($P(\hat{Y}=1|Y=0)$) to be equal:
\begin{equation}%\small
    P(\hat{Y}{=}1 \mid Y{=}y, Z{=}0)=P(\hat{Y}{=}1 \mid Y{=}y, Z{=}0) \; . \label{eq:equalizedodds}
\end{equation}
The corresponding counterfactual utility definition is
\begin{equation}%\small
    \begin{aligned}
        &\bigl( P(W_m \geq \tau \mid \Gamma{=}1, Z{=}0) = P(W_m \geq \tau \mid \Gamma{=}1, Z{=}1) \bigr) \\ \label{equalizedodds_cfutil_literal}
        \land &\bigl( P(W_m \geq \tau \mid \Gamma{=}0, Z{=}0) = P(W_m \geq \tau \mid \Gamma{=}0, Z{=}1) \bigr) \; .
    \end{aligned}
\end{equation}
% In the binary classification setting, Equation \ref{eq:equalizedodds} is actually the same as \texttt{EqOppCfUtil} (Equation \ref{eq:fdmp_eqopp}) with the thresholds removed:
% \begin{equation}
%     P(W_m \mid Z=0)=P(W_m \mid Z=1) \; . \label{eq:equalizedodds_cfutil_nothresholds}
% \end{equation}
% We discuss removing thresholds further in Appendix \ref{appendix:threshold_alternatives}.

\paragraph{Predictive Parity}
\textit{Predictive Parity} \cite{chouldechova2017fair} is essentially the inverse of Equal Opportunity, which requires that the probability that an individual predicted to be positive actually belongs to the positive class is equal for both groups:
\begin{equation}%\small
    P(Y{=}1 \mid \hat{Y}{=}1, Z{=}0)=P(Y{=}1 \mid \hat{Y}{=}1, Z{=}1) \; \label{eq:predictive_parity}
\end{equation}
The respective counterfactual utility definition is
\begin{equation}%\small
    P(\Gamma{=}1 \mid W\geq\tau,Z{=}0)=P(\Gamma{=}1 \mid W\geq\tau,Z{=}1) \; . \label{eq:predictive_parity_cfutil}
\end{equation}

\paragraph{Conditional Demographic Parity}
\textit{Conditional Demographic Parity}~\cite{corbett2017algorithmic} extends Demographic Parity (Definition \ref{def:dem_parity}) by allowing one or more legitimate attributes $L$ to impact the outcome of the decision:
\begin{equation}%\small
    P(\hat{Y}{=}1 \mid L{=}l,Z{=}0) = P(\hat{Y}{=}1 \mid L{=}l,Z{=}1) \; .\label{eq:conditional_demparity}
\end{equation}
for some $l$.
In our framework, this is:
\begin{equation}%\small
    P(W_m \geq \tau \mid L{=}l, Z{=}0)=P(W_m \geq \tau \mid L{=}l, Z{=}1) \; . \label{eq:conditional_demparity_cfutil}
\end{equation}
Here $L$ is playing a similar role as $\Gamma$ in Equation \ref{eq:fdmp_eqopp_gamma}, as it requires equal welfare for some subset of the general population.

\paragraph{Predictive Equality}
\textit{Predictive Equality} ~\cite{chouldechova2017fair} is satisfied if individuals in the negative class have equal probabilities of receiving a positive prediction for each protected group:
\begin{equation}%\small
    P(\hat{Y}{=}1 \mid Y{=}0, Z{=}0) = P(\hat{Y}{=}1 \mid Y{=}0, Z{=}1) \label{eq:predictive_equality}
\end{equation}
which, in our framework, translates to
\begin{equation}%\small
    P(W_m \geq \tau \mid \Gamma{=}0, Z{=}0) = P(W_m \geq \tau \mid \Gamma{=}0, Z{=}1) \; . \label{eq:predictive_equality_cfutil}
\end{equation}

\paragraph{Conditional Use Accuracy Equality}
\textit{Conditional Use Accuracy Equality} ~\cite{berk2018fairness} requires the probability for individuals with positive predictions to belong to the positive class to be equal for both protected groups, \textit{and} the probability for individuals with negative predictions to belong to the negative class to be equal for both protected groups:
{%\small
\begin{align}
    &\bigl( P(Y{=}1 \mid \hat{Y}{=}1, Z{=}0) = P(Y{=}1 \mid \hat{Y}{=}1, Z{=}1) \bigr) \\ \label{eq:conditional_use_accuracy}
    \land &\bigl( P(Y{=}0 \mid \hat{Y}{=}0, Z{=}0) = P(Y{=}0 \mid \hat{Y}{=}0, Z{=}1) \bigr)  \nonumber
\end{align}}%
The counterfactual utility equivalent is:
{%\small
\begin{align}
    &\bigl( P(\Gamma{=}1 \mid W_m\geq\tau, Z{=}0) = P(\Gamma{=}1 \mid W_m\geq\tau, Z{=}1) \bigr) \\ \label{eq:conditional_use_accuracy_cfutil}
    \land &\bigl( P(\Gamma{=}0 \mid W_m < \tau, Z{=}0) = P(\Gamma{=}0 \mid W_m < \tau, Z{=}1) \bigr) \; .  \nonumber
\end{align}}%

\paragraph{Overall Accuracy Equality}
\textit{Overall Accuracy Equality} ~\cite{berk2018fairness} requires the probability that an individual is assigned to their true class to be equal for both protected groups:
\begin{equation}%\small
    P(\hat{Y}{=}Y, Z{=}0) = P(\hat{Y}{=}Y, Z{=}1) \label{eq:overall_accuracy}
\end{equation}
We interpret $Y=1$ as $\Gamma=1$ and $\hat{Y}=1$ as $W_m \geq \tau$.
Similarly, we interpret $Y=0$ as $\Gamma=0$ and $\hat{Y}=0$ as $W_m \geq \tau$.
So Equation \ref{eq:overall_accuracy} translates in our framework as:
\begin{equation}%\small
    \begin{aligned}
        &\bigl( P(W_m\geq\tau, \Gamma{=}1, Z{=}0)  = P(W_m\geq\tau, \Gamma{=}1, Z{=}1) \bigr) \\ \label{eq:overall_accuracy_cfutil}
        \land &\bigl( P(W_m < \tau, \Gamma{=}0, Z{=}0)  = P(W_m < \tau, \Gamma{=}0, Z{=}1) \bigr) \; .
    \end{aligned}
\end{equation}

\paragraph{Treatment Equality}
\textit{Treatment Equality}~\cite{berk2018fairness} requires an equal ratio of false negatives ($P(\hat{Y}{=}0|Y{=}1)$) and false positives ($P(\hat{Y}{=}1|Y{=}0)$) for each protected group:
\begin{equation}%\small
    \frac{P(\hat{Y}{=}0 \mid Y{=}1, Z{=}0)}{P(\hat{Y}{=}1 \mid Y{=}0, Z{=}0)} = \frac{P(\hat{Y}{=}0 \mid Y{=}1, Z{=}1)}{P(\hat{Y}{=}1 \mid Y{=}0, Z{=}1)} \label{eq:treatment_equality}
\end{equation}
which translates to our framework as:
\begin{equation}%\small
    \frac{P(W_m < \tau \mid \Gamma{=}1, Z{=}0)}{P(W_m \geq \tau \mid \Gamma{=}0, Z{=}0)} = \frac{P(W_m < \tau \mid \Gamma{=}1, Z{=}1)}{P(W_m \geq \tau \mid \Gamma{=}0, Z{=}1)}  \label{eq:treatment_equality_cfutil} \; .
\end{equation}

\paragraph{Test Fairness}
\textit{Test Fairness}~\cite{chouldechova2017fair} applies to classifiers that predict a probability $S$ rather than a binary class $Y$.
A classifier satisfies Test Fairness if, for any predicted probability $S$, individuals in each protected group have equal probability of being in the positive class:
\begin{equation}%\small
    P(Y{=}1 \mid S{=}s, Z{=}0) = P(Y{=}1 \mid S{=}s, Z{=}1) \; \forall s \in [0,1] \; . \label{eq:test_fairness}
\end{equation}
We can implement this in our counterfactual utility framework as
\begin{equation}%\small
    P(\Gamma{=}1 \mid W_m{=}w, Z{=}0) = P(\Gamma{=}1 \mid W_m{=}w, Z{=}1) \; \forall w \in \mathbb{R} \; . \label{eq:test_fairness_cfutil}
\end{equation}

% \section{Design Choices and Alternatives}\label{appendix:design_choices_and_extensions}
\subsection{Alternatives to Thresholds.}
\label{appendix:threshold_alternatives}
Our choice to use thresholds $\tau$ and $\rho$ when defining \texttt{DemParWelf} and \texttt{EqOppCfUtil} is not the only option.
Intuitively, what is needed is some way to compare the distributions $P(W|Z{=}0)$ and $P(W|Z{=}1)$.
Our use of thresholds reduces this comparison to a simple binary, as with the traditional definitions.
We chose this for its simplicity and compatibility, as it easily allows standard techniques like quantifying the extent to which it is satisfied by computing $P(W_m \geq \tau|Z{=}0)-P(W_m \geq \tau|Z{=}1)$.  
Instead, we could have defined Demographic Parity as
\begin{equation}%\small
    P(W_m \mid Z{=}0)=P(W_m \mid Z{=}1) \; . \label{eq:fdmp_dempar_alt_strict}
\end{equation}
This is a much stricter constraint as it forces equality of the entire distribution, and may even be impossible to enforce for some welfare functions.
Or, perhaps more practically, we could compare the expected utilities of the two groups:
\begin{equation}%\small
    \mathbb{E}[W_m \mid Z{=}0] = \mathbb{E}[W_m \mid Z{=}1] \; . \label{eq:fdmp_dempar_alt_expected}
\end{equation}
Depending on the way utilities are defined and thresholds are chosen in a specific problem, this could be equivalent to, stricter than, or simply different from our version.
% For instance, if the welfare distributions are heavily skewed such that $\mathbb{E}[W_m]$ is overly influenced by outliers, then a threshold is likely the better option.

The choice may also depend on the notion of fairness for the problem.
For instance, in the two-stage loan application MDP example in
%Section \ref{subsection:applications_objective_not_predict_target} and
Appendix \ref{appendix:worked_example_two_stage_loan_mdp}, we define fairness by requiring equal probability of worst-case scenarios across both groups.
Alternatively, consider a decision algorithm that determines the salaries for each employee of a large corporation.
In this scenario, equal expected welfare across protected groups (Equation \ref{eq:fdmp_dempar_alt_expected}) could be satisfied by giving a small number of high-ranking minority employees a very large salary, but paying all low-ranking minority employees smaller salaries than their majority counterparts.
A salary threshold (Equation \ref{eq:fdmp_dempar}) will not work either, since it will only enforce equal ratios of employees above a certain level, rather than ensure equal pay for employees at the same level.
In order to enforce equal pay at each level within the organization, we would want to use a form of Equation \ref{eq:fdmp_dempar_alt_strict} that requires equal welfare distributions.
As this example suggests, there are still other summary statistics that could be considered.
For example, the median welfare would have a similar robustness to outlier individuals and indeed our use of thresholds corresponds to the value of the CDF at a particular point.

\subsection{More than Two Protected Groups}\label{appendix:more_than_two_protected groups}
For ease of exposition, this work discusses only the case when there are two protected groups.
However, our insights and contributions are focused on assumptions about parameters other than $Z$, and therefore extend naturally to cases with multiple protected attributes.
Specifically, when there are more than two protected groups ($|\mathcal{Z}|>2$), \texttt{DemParWelf} becomes:
\begin{equation}%\small
    \begin{aligned}
        P (W_m\geq \tau \mid Z{=}0 ) = P (W_m\geq \tau \mid Z{=}z ) \\
        \forall z\in\{1,2,...,|\mathcal{Z}|-1\} \label{eq:fdmp_dempar_z>2}
    \end{aligned}
\end{equation}
and \texttt{EqOppCfUtil} becomes:
\begin{equation}%\small
    \begin{aligned}
        P (W_m\geq \tau \mid \Gamma{=}1, Z{=}0 ) = P (W_m\geq \tau \mid \Gamma{=}1, Z{=}z ) \\
        \forall z\in\{1,2,...,|\mathcal{Z}|-1\} \; . \label{eq:fdmp_eqopp_z>2}
    \end{aligned}
\end{equation}

% \section{Fully Worked Examples}\label{subsection:appendix_worked_examples}
\subsection{Fully Worked Examples}\label{subsection:appendix_worked_examples}

\subsubsection{Self-Fulfilling Prophecies}\label{appendix:worked_example_recidivism_prediction}
Here we provide a more complete example of the recidivism prediction problem from Section \ref{subsection:applications_pred_infl_outcomes}, which
illustrates how Assumption \ref{assumption:decision_influence_outcome} allows for self-fulfilling prophecies with \texttt{EqOppClf} but no0t with \texttt{EqOppCfUtil}.
Additionally, we show that \textit{principal fairness}~\cite{imai2020principal}, which also prevents self-fulfilling prophecies, is a special case of our framework.

We use the recidivism prediction example posed by Imai and Jiang~\cite{imai2020principal} where a binary classifier predicts whether an inmate convicted of a crime will recidivate.
The target variable $Y$ corresponds to whether or not the inmate will recidivate, with $Y=0$ indicating recidivism.
This problem differs from typical classification since $Y$ is influenced by $\hat{Y}$.
When decisions influence the observed target variable, it is helpful to visualize the dataset by \textit{principal strata} ~\cite{frangakis2002principal} where each principal stratum characterizes how an individual would be affected by the decision $\hat{Y}$ with respect to the variable of interest $Y$. %~\cite{imai2020principal}.
Since this is a binary classification problem with binary decisions and binary targets, we have a total of four principal strata.
We assign labels to each stratum according to their behavior in Table \ref{tab:recidivism_stratum}.
For example, an individual in the \texttt{Backlash} stratum will recidivate if they are detained, so $P(Y{=}1 \mid \hat{Y}{=}0)=0$, but will not recidivate if released, so $P(Y{=}1 \mid \hat{Y}{=}1)=1$.%
To model inmates who always prefer to be released, we can take the welfare function to be
a binary function with $W=1$ when the inmate is released and $W=0$ when detained.
Similarly, to model a judge (decision-maker) who always prefers outcomes where the inmate does not recidivate, we can set the cost function to $C=1$ when the inmate recidivates and $C=0$ when they do not.

Following a similar fairness criteria of that posed by \cite{imai2020principal}, we want to ensure that inmates who will not recidivate if released are released with equal probability for each protected group.
Therefore, we set the welfare threshold $\tau=1$ so that a good outcome from an inmate's perspective is when they are released.
Similarly, we set the cost threshold $\rho=0$ such that a good outcome from the judge's perspective is when an inmate does not recidivate.

% % Assign FDMP parameters
% In order to leverage our counterfactual utility framework, we need to establish FDMP parameters $(I, Z, M, W, C, \tau, \rho)$.
%     The individual's unprotected attributes $I=(X,Y^P)$.
%     The decision-algorithm space $M=(\mathcal{X}\times \mathcal{Z})\rightarrow \mathcal{Y}\;$.
%     We define the welfare function to be $W=1$ if the inmate is released ($\hat{Y}=1$) and $W=0$ if detained ($\hat{Y}=0$).
%     We define the cost function to be $C=1$ if the inmate recidivates and $C=0$ if they do not.
%     We consider a good outcome for the inmate to be when they are released ($\tau=1$),\footnote{
%         One might argue that a preventable individual would be better off being detained since they will not get into future trouble.
%         However, because this is a one-step decision problem, future timesteps are not considered.
%     }
%     and a good outcome for the judge (decision-maker) to be when the inmate does not recidivate ($\rho=0$).
We wish to evaluate the fairness of a classifier $\hat{Y}^\dagger$ that produces the results shown in Table \ref{tab:recidivism_stratum_full}.
We compare three different fairness definitions when evaluating $\hat{Y}^\dagger$: \texttt{EqOppClf}, \texttt{EqOppCfUtil}, and principal fairness. 

% \paragraph{Evaluating \texttt{EqOppClf}}
An individual is qualified under \texttt{EqOppClf} if $Y$ is observed to be $1$.
This means that an inmate is qualified if  they do not recidivate, which corresponds to inmates that (a) are in the \texttt{Safe} stratum, (b) are in the \texttt{Backlash} stratum and are released, or (c) are in the \texttt{Preventable} stratum and are detained.
Thus, \texttt{EqOppClf} becomes:
\begin{equation*}%\small
    \begin{aligned}
        P\bigl(\hat{Y}{=}1 \bigm| Z{=}0, &(Y^P=\text{Backlash}\land \hat{Y}{=}1) \\
                                         &\lor (Y^P=\text{Preventable}\land \hat{Y}{=}0) \\
                                         &\lor (Y^P=\text{Safe}) \bigr) \\
        = P\bigl(\hat{Y}{=}1 \bigm| Z{=}1, &(Y^P=\text{Backlash}\land \hat{Y}{=}1) \\
                                                         &\lor (Y^P=\text{Preventable}\land \hat{Y}{=}0) \\
                                                         &\lor (Y^P=\text{Safe}) \bigr) \; .
    \end{aligned}
\end{equation*}
Therefore, even if the minority and majority inmate populations are identical in every way other than their protected attribute, a classifier could satisfy \texttt{EqOppClf} while having different release rates for inmates who would not recidivate.
This can be done through a self-fulfilling prophecy where the classifier manipulates who is considered "qualified".
$\hat{Y}^\dagger$ accomplishes this by detaining more \texttt{Backlash} minority inmates and releasing more \texttt{Preventable} minority inmates (Table \ref{tab:recidivism_stratum_full}), while still satisfying \texttt{EqOppClf}:
\begin{equation*}%\small
    \begin{aligned}
        \frac{20+160}{20+80+40+160} \stackrel{?}{=}& \frac{20+160}{20+80+40+160} \\
        \frac{3}{5} =& \frac{3}{5} \;  .
    \end{aligned}
\end{equation*}
$\hat{Y}^\dagger$ causes two-thirds of the minority ($Z=0$) Backlash inmates to recidivate by detaining them, thus rendering them unqualified according to \texttt{EqOppClf}.~\footnote{
    Similarly, the detained Preventable inmates were "manipulated" into not recidivating.}
Since $\hat{Y}^\dagger$ detains only half of the majority ($Z=1$) Backlash inmates, this results in a larger proportion of minority inmates who were rendered unqualified through detainment.
This results in a self-fulfilling prophecy since $\hat{Y}^\dagger$ satisfies \texttt{EqOppClf} by biasing the selection of qualified inmates rather than by making fair decisions.
% since the only reason the Backlash inmates are considered unqualified, and therefore not included in the \texttt{EqOppClf} computation, is because they were detained, which is a poor decision for both the inmate and the decision-algorithm.

% Compute EqOppCfUtil
% \paragraph{Evaluating \texttt{EqOppCfUtil}}
Conversely, \texttt{EqOppCfUtil} does not allow for self-fulfilling prophecies since it has a prediction-independent definition of "qualification".
Referencing Equation \ref{eq:fdmp_eqopp_gamma}, an individual is qualified under \texttt{EqOppCfUtil} with $\tau=1$ and $\rho=0$ if $\exists \hat{Y'} \in \hat{\mathcal{Y}} : W_{\hat{Y}'} \geq 1 \land C_{\hat{Y}'} \leq 0\;$.
In other words, an inmate is considered qualified if there exists a classifier that will produce $W\geq 1$ and $C\leq 0$, which is only possible for individuals who will not recidivate when released.
Thus, according to \texttt{EqOppCfUtil}, an inmate is qualified if they are in the \texttt{Safe} or \texttt{Backlash} stratum, regardless of if they are detained or released.
\texttt{EqOppCfUtil} is then evaluated as:
\begin{equation*}%\small
    \begin{aligned}
         &P \bigl( \hat{Y}{=}1 \bigm| Z{=}0, Y^P\in\{\text{Safe},\text{Backlash}\} \bigr) \\
         \stackrel{?}{=} &P \bigl( \hat{Y}{=}1 \bigm| Z{=}1, Y^P\in\{\text{Safe},\text{Backlash}\} \bigr) \\
    \end{aligned}
\end{equation*}
\begin{equation*}%\small
    \begin{aligned}
        \frac{20+160}{40+20+40+160} &\stackrel{?}{=} \frac{20+160}{20+20+40+160} \\
        % 180/260 &= 180/240 \\
        % \frac{9}{13} &< \frac{3}{4} \; .
        .692 &< .750 \; .
    \end{aligned}
\end{equation*}
As expected, the proportion of qualified minority inmates who were released ($.692$) is less than that of majority inmates ($.750$), which means that $\hat{Y}^\dagger$ does not satisfy \texttt{EqOppCfUtil}.
% Explain why EqOppClf is better.
Contrasted against \texttt{EqOppClf} which requires equal release rates for those \textit{observed} to not recidivate, \texttt{EqOppCfUtil} accounts for counterfactuals by requiring equal release rates for those who would not recidivate \textit{if released}.
By considering counterfactuals, \texttt{EqOppCfUtil} ensures fairness is not satisfied through self-fulfilling prophecies.

% Define principal fairness
% \paragraph{Evaluating Principal Fairness}
Although it is a stricter set of requirements than Equal Opportunity, principal fairness~\cite{imai2020principal} also aims to prevent self-fulfilling prophecies by requiring equal release rates for each principal stratum.
To demonstrate the robustness of our FDMP model, we will implement principal fairness as a FDMP instantiation.
If there are $p$ principal strata, principal fairness is defined as a conjunction of $p$ constraints:
% Define Principal Fairness in our framework
\begin{equation}%\small
    \begin{aligned}
        P(W_C\geq\tau \mid Z{=}0,\Gamma^i{=}1) = P(W_C\geq\tau \mid Z{=}1,\Gamma^i{=}1) \\
        \forall i\in \{0,...,p-1\} \label{eq:special_case_princ_fair}
    \end{aligned}
\end{equation}
where $\Gamma^i=1$ if the individual is in the $i^\text{th}$ principal stratum.
% $Y^P=Y^{P=i}$.
% Compute Principal Fairness in our framework
For the recidivism prediction problem, $W_C=W=\hat{Y}$, so Equation \ref{eq:special_case_princ_fair} corresponds to
\begin{subequations}%\small
    \begin{alignat}{1}
        &P(\hat{Y}\geq 1 \mid Z{=}0, \text{Danger}) 
            = P(\hat{Y}\geq 1 \mid Z{=}1, \text{Danger}) \label{eq:princ_fair_dangerous} \\
        \land \; &P(\hat{Y}\geq 1 \mid Z{=}0, \text{Backlash}) 
            = P(\hat{Y}\geq 1 \mid Z{=}1, \text{Backlash}) \label{eq:princ_fair_backlash} \\
        \land \; &P(\hat{Y}\geq 1 \mid Z{=}0, \text{Prevent}) 
            = P(\hat{Y}\geq 1 \mid Z{=}1, \text{Prevent}) \label{eq:princ_fair_preventable} \\
        \land \; &P(\hat{Y}\geq 1 \mid Z{=}0, \text{Safe}) 
            = P(\hat{Y}\geq 1 \mid Z{=}1, \text{Safe}) \label{eq:princ_fair_safe} \; .
    \end{alignat}
\end{subequations}
Principal fairness is not satisfied by $\hat{Y}^\dagger$ since the release rates of the Backlash (\ref{eq:princ_fair_backlash}) and Preventable (\ref{eq:princ_fair_preventable}) strata are unequal between protected groups.
I.e. for Backlash:
\begin{equation*}%\small
    \begin{aligned}
        &P(\hat{Y}\geq 1 \mid Z{=}0, Y^P=\text{Backlash}) \\
            &\stackrel{?}{=} P(\hat{Y}\geq 1 \mid Z{=}1, Y^P=\text{Backlash}) \\
    \end{aligned}
\end{equation*}
\begin{equation*}%\small
    \begin{aligned}
        \frac{20}{40+20} &\stackrel{?}{=} \frac{20}{20+20} \\
        \frac{1}{3} &\neq \frac{1}{2} \; .
    \end{aligned}
\end{equation*}
Generally, we prefer \texttt{EqOppCfUtil} over principal fairness since the latter requires equality in strata that may be irrelevant to fairness (e.g. requiring equal release rates in the Dangerous strata is not relevant since releasing a Dangerous inmate is an undesirable outcome).
However, as demonstrated, principal fairness is well defined within our FDMP model.

\begin{table*}
    %\small
    \begin{center}
    \begin{tabular}{c c c | c c c}
        \multicolumn{3}{c|}{$Z=0$} & \multicolumn{3}{|c}{$Z=1$} \\ \hline
        & Dangerous & Backlash & & Dangerous & Backlash \\
        Detained & 120 & 40 & Detained & 80 & 20 \\
        Released & 30  & 20 & Released & 20  & 20 \\ \hline
        & Preventable & Safe & & Preventable & Safe \\
        Detained & 80 & 40 & Detained & 80 & 40 \\ 
        Released & 10  & 160 & Released & 80  & 160
    \end{tabular}
    \end{center}
    \caption{
        A numerical illustration of the results of the predictions from classifier $\hat{Y}^\dagger$ on 1,000 inmates, separated by protected attribute and principal stratum.
        Each cell represents the number of inmates in the principal stratum and protected group who were detained ($\hat{Y}=0$) and released ($\hat{Y}=1$).
        The table is partially reproduced from Imai and Jiang's example~\cite{imai2020principal} which represents the results of a classifier that 
        satisfies \texttt{EqOppClf}, \texttt{EqOppCfUtil}, and principal fairness.
        % does not satisfy \texttt{EqOppClf} but does satisfy principal fairness.
        However, we modified the $Z=0$ numerical results to demonstrate a scenario where \texttt{EqOppClf} is satisfied, but \texttt{EqOppCfUtil} and principal fairness are not.
        See Table \ref{tab:recidivism_stratum} for the definitions of each principal stratum. 
    }
    \label{tab:recidivism_stratum_full}
\end{table*}

% \subsection{Fairness in Reinforcement Learning}\label{appendix:worked_example_two_stage_loan_mdp}
\subsubsection{Fairness in Reinforcement Learning}\label{appendix:worked_example_two_stage_loan_mdp}

Here we continue the discussion from Section \ref{subsection:applications_objective_not_predict_target} on applying our framework to reinforcement learning, and focus the discussion on a two-stage loan application MDP.
We apply Equations \ref{eq:cost_function_rl} and \ref{eq:welfare_function_rl} to Equation \ref{eq:fdmp_eqopp} and \ref{eq:fdmp_eqopp_gamma} in order to define \texttt{EqOppCfUtil} in the RL setting.
We then compare \texttt{EqOppCfUtil} to that of \cite{wen2021algorithms}
%which also provides an RL-translation of Equal Opportunity using an individual utility function, but does not fully consider counterfactual scenarios.
who also provide an RL-translation of Equal Opportunity using an individual utility function, but do not fully consider counterfactual scenarios.%
\footnote{They also provide an RL-translation of demographic parity.  As this does not involve qualification, it is equivalent to \texttt{DemParWelf} in this setting.  They do not examine the ability of this approach to extend to non-RL settings.}

\begin{figure*}
    \centering
    \frame{\includegraphics[scale=1]{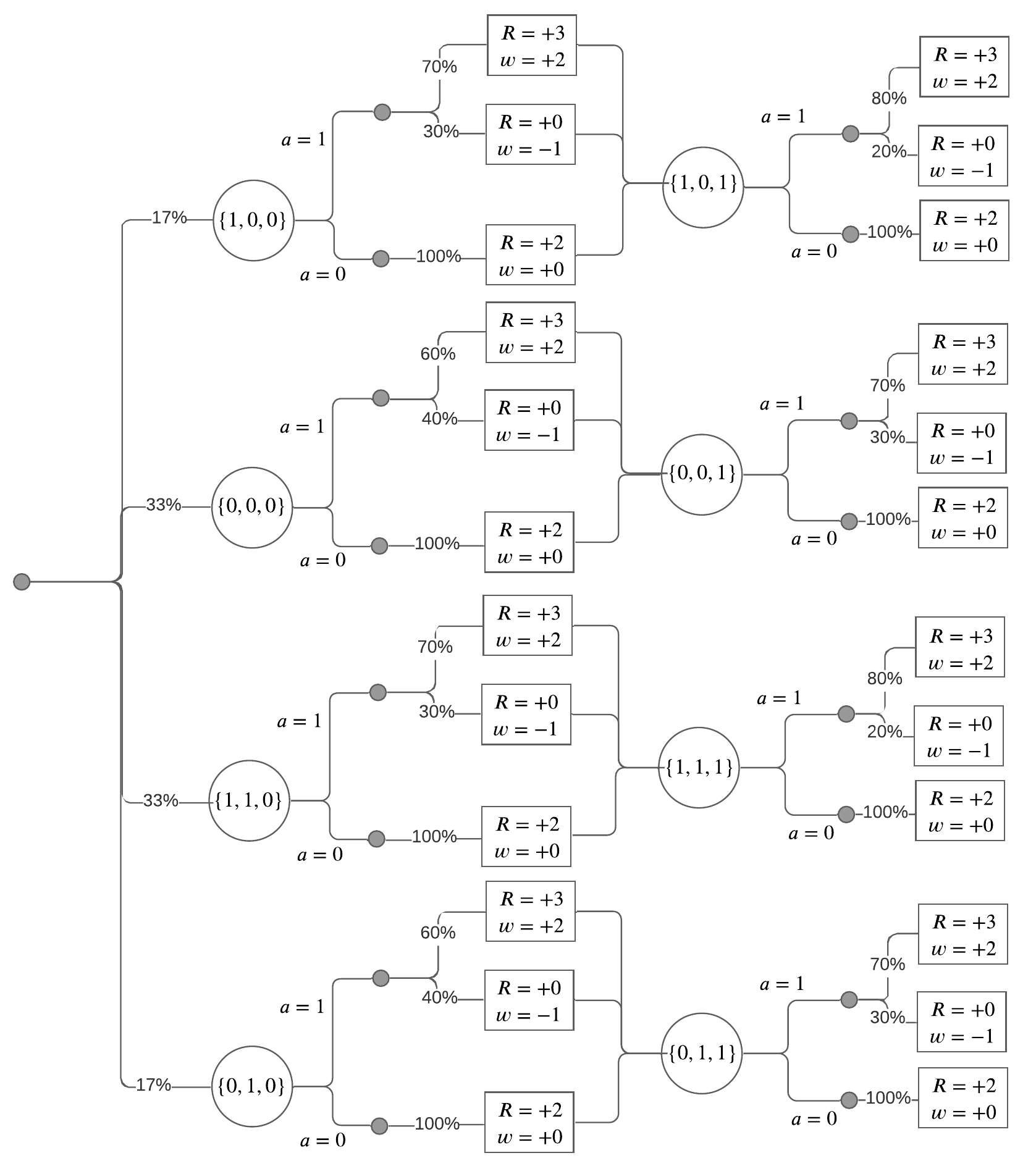}}
    \caption{
        Two-stage loan application MDP with two actions and eight total states.
        Welfare contributions $w(s,a)$ are displayed alongside the rewards $R(s,a)$.
        States (large circles) have 3 parameters, with the first element indicating if the applicant is prime (1) or subprime (0); the second element is the binary protected attribute $Z$ with $Z=0$ indicating the minority and $Z=1$ the majority; and the third element is the loan timestep with zero indicating the first timestep and 1 indicating the second timestep.
        The action $a=1$ corresponds to the lender granting the applicant a loan, and $a=0$ corresponds to the lender rejecting the applicant.
        The four left-most percentages represent the initial state distribution $\mu$; e.g. an applicant sampled from $\mu$ has a 17\% probability of being a minority, subprime applicant.
        The remaining percentages represent the joint probabilities of the welfare contributions $w$ and rewards $R$ occurring, given the selected action.
    }
    \label{fig:two_stage_loan_mdp}
\end{figure*}

As a motivating example, we consider a two-stage loan application decision process represented as an MDP, where a loan applicant applies for loans in two sequential timesteps, as shown in Figure \ref{fig:two_stage_loan_mdp}.
The decision-maker corresponds to the lender, who is represented by a policy which can either grant or reject the applicant's loan application in each timestep.
There are two types of applicants.
    The first type, \textit{prime}, will pay back a loan with 70\% probability in the first timestep, and 80\% in the second timestep.
    The second type, \textit{subprime}, will pay back a loan with 60\% probability in the first, and 70\% in the second.
Applicants in the minority group are twice as likely to be subprime as prime, whereas applicants in the majority group are twice as likely to be prime as subprime.
% An applicant is equally likely to be in the minority as the majority.
The MDP state includes the applicant's behavior type (prime or subprime), protected attribute (minority or majority), and the timestep (0 or 1).
The reward function $R$ is defined so that the lender benefits when a loan is repaid, loses when a loan is defaulted on, and is indifferent when a loan is rejected.
Table \ref{tab:appendix_two_stage_loan_welfare_cost} provides the full definition of $R$.

% \paragraph{Two-stage loan FDMP}
FDMP parameters $I$, $Z$, $C$, and $M$ can be inferred from the MDP as described in Section \ref{subsection:applications_objective_not_predict_target}, so we only need to define $W$, $\tau$, and $\rho$.
% Since the state is fully observable to the policy, the policy space $M=S\rightarrow\mathcal{A}=\tilde{S}\times\mathcal{Z}\rightarrow\mathcal{A}$.
% The welfare function $W$ is defined as the expected sum of welfare contributions $w(s,a)$ across both timesteps.
% Similarly, the cost function $C$ is the expected sum of negative rewards over both timesteps.\footnote{
%     For clarity we shift the cost so that there are no negative values: $C=-R+3$.}
Next, we need to define the welfare contribution function $w$, from which we can construct the welfare function $W$.
Similar to the single-stage loan example in Section \ref{subsection:applications_preds_dont_imply_eq_outcomes}, we define the the welfare contribution function $w$ so that an applicant benefits when they repay a loan, loses when they default on a loan, and is indifferent when rejected.
The full definition of $w$ is defined in Table \ref{tab:appendix_two_stage_loan_welfare_cost}.

\begin{table*}[htb]
    \small
    \center
    \begin{tabular}{c c c|c|c|c|c|c}
        \multicolumn{3}{c|}{State $s$} & Action $a$ & Outcome & $w$ & $R$ & Probability \\
        Applicant Type & $Z$ & Timestep & & & & & \\ \hline
        \multirow{2}{*}{Prime}    & \multirow{2}{*}{*} & \multirow{2}{*}{0} & \multirow{2}{*}{Grant}  & Repaid    & +2 & +3 & .7 \\ 
                                  &                    &                    &                         & Defaulted & -1 &  0 & .3 \\ \hline
        \multirow{2}{*}{Subprime} & \multirow{2}{*}{*} & \multirow{2}{*}{0} & \multirow{2}{*}{Grant}  & Repaid    & +2 & +3 & .6 \\ 
                                  &                    &                    &                         & Defaulted & -1 &  0 & .4 \\ \hline
        \multirow{2}{*}{Prime}    & \multirow{2}{*}{*} & \multirow{2}{*}{1} & \multirow{2}{*}{Grant}  & Repaid    & +2 & +3 & .8 \\ 
                                  &                    &                    &                         & Defaulted & -1 &  0 & .2 \\ \hline
        \multirow{2}{*}{Subprime} & \multirow{2}{*}{*} & \multirow{2}{*}{1} & \multirow{2}{*}{Grant}  & Repaid    & +2 & +3 & .7 \\ 
                                  &                    &                    &                         & Defaulted & -1 &  0 & .3 \\ \hline
        \multirow{2}{*}{*}        & \multirow{2}{*}{*} & \multirow{2}{*}{*} & \multirow{2}{*}{Reject} & \multirow{2}{*}{Rejected} & \multirow{2}{*}{0} &  \multirow{2}{*}{+2} & \multirow{2}{*}{1} \\
                                  &                    &                    &                         &           &    &    &    \\
    \end{tabular}
    \caption{
        Joint distributions for the welfare contributions $w$ and rewards $R$ for the two-stage loan MDP example.
        The * character serves as a wildcard and represents "any value".
        As an example of interpreting this table, the first row is interpreted as follows: a Prime applicant in the initial timestep that is granted a loan will repay the loan with 70\% probability, which yields $w=+2$ and $R=+3$, and will default on the loan with 30\% probability, yielding $w=-1$ and $R=0$.
        These values are also shown within the MDP diagram in Figure \ref{fig:two_stage_loan_mdp}.
    }
    \label{tab:appendix_two_stage_loan_welfare_cost}
\end{table*}

In order to define the remaining FDMP parameters $\tau$ and $\rho$, we first need to establish our fairness objective.
Building on the \textit{no unnecessary harm} principle \cite{ustun2019fairness,martinez2020minimax}, we aim to to ensure that the lender does not cause significant harm to one protected group more than the other, unless doing so avoids severe harm to the lender.~\footnote{
    Our meaning of the \textit{no unnecessary harm} principle is slightly different from other works.
    For example, \cite{ustun2019fairness, martinez2020minimax} use it to mean that one protected group's welfare should not decrease unless it increases the welfare of another protected group.
    Here, we use it to mean that the probability difference of causing negative welfare between the protected groups should not increase unless doing so decreases the cost for the decision-maker.
    % In other words, other works consider the principle to mean Pareto optimality across all protected groups' welfare functions, whereas we consider it as Pareto optimality across the cost function and the probability difference in producing negative welfare between the protected groups.
    Intuitively, other works consider the principle to mean Pareto optimality across all of the protected groups, whereas we consider it as Pareto optimality across all protected groups plus the decision-maker.
    }
We consider significant harm for the applicant to be when they default twice, when they are rejected twice, or when they are rejected once and default once.
In other words, we consider a policy to be causing an applicant significant harm unless at least one loan is granted and repaid.
This corresponds to a welfare threshold of $\tau=1$.
% -4 is computed by finding the minimum reward possible that has no defaults, then multiplying by -1. This corresponds to two rejections which have R=4 combined.
From the lender's perspective, we consider significant harm to be when the applicant defaults at least once, which corresponds to a cost threshold of $\rho=-4$.
% Following the the cost matrix in \cite{Dua:2019}, we consider a catastrophic outcome for the lender to be when one or more defaults occur, corresponding to $\rho=2$.
Because our fairness objective considers both the applicant's welfare and lender's cost, we will want to use Equal Opportunity as our fairness definition.

Suppose we are given the policy $\pi^\text{Prime}$ that assigns loans to all prime applicants and rejects loans to all subprime applicants, and we wish to evaluate if $\pi^\text{Prime}$ satisfies our fairness objective.
First, we will evaluate Wen, Bastani, and Topcu's MDP translation of Equal Opportunity~\cite{wen2021algorithms}, hereafter referred to as \texttt{EqOppMDPStatic}.
%, which they also apply to a repeat-loan application MDP.
\texttt{EqOppMDPStatic} requires the cumulative expected individual rewards (welfare) to be equal for qualified individuals in both protected groups:
\begin{equation*}%\small
    \mathbb{E}(W_{\pi} \mid p_0\geq \alpha, Z{=}0) = \mathbb{E}(W_{\pi} \mid p_0\geq \alpha, Z{=}1)
\end{equation*}
where $p_0$ is the individual's probability of repaying the loan in the first timestep, and $\alpha$ is some qualification threshold.
We can apply \texttt{EqOppMDPStatic} to our two-stage loan MDP example by selecting the qualification threshold $\alpha$, which we set to $\alpha=2/3$ since the optimal policy grants loans to applicants with a repayment probability of at least $2/3$.
This means prime applicants ($p_0=.7$) are qualified under \texttt{EqOppMDPStatic} while subprime applicants ($p_0=.6$) are not.
Therefore, the policy $\pi^{\text{Prime}}$ that grants loans to all prime applicants and rejects all loans to subprime applicants is fair according to \texttt{EqOppMDPStatic} since
\begin{align*}
    \mathbb{E}(W_{\pi^{\text{Prime}}} \mid \text{Prime}, Z{=}0) = \mathbb{E}(W_{\pi^{\text{Prime}}} \mid \text{Prime}, Z{=}1) \;.
\end{align*}

However, subprime applicants are as likely to repay a loan in the second timestep as prime applicants are in the first.
Certainly the lender would prefer to grant them loans, so it seems unfair to say that subprime applicants are forever unqualified just because they are initially beneath the qualification threshold.
Instead, we want our fairness definition to be able to understand that qualification may be a moving target, and that the applicants' repayment probability in later timesteps should also be considered.

% \paragraph{Evaluating \texttt{EqOppCfUtil}}
\texttt{EqOppCfUtil}, on the other hand, considers an applicant to be qualified if there exists a policy that will result in good outcomes for both the applicant and the lender:
\begin{equation}%\small
    \begin{aligned}
    P(W_{\pi^{\text{Prime}}} \geq \tau \mid \Gamma{=}1, Z{=}0) = P(W_{\pi^{\text{Prime}}} \geq \tau \mid \Gamma{=}1, Z{=}1)
    \end{aligned}
    \label{eq:eqoppcfutil_rl}
\end{equation}
where $\Gamma$ is an indicator variable representing qualified individuals:
\begin{equation*}%\small
    \Gamma=\begin{cases}
        1 \; \text{ if } \; \exists \pi' \in \Pi : W_{\pi'} \geq \tau \land C_{\pi'} \leq \rho \\
        0 \; \text{ otherwise} \; .
    \end{cases}
\end{equation*}
This means that qualification under \texttt{EqOppCfUtil} is determined by the applicant's repayment probability across both timesteps, rather than just the initial timestep as in \texttt{EqOppMDPStatic}. 
When we initialized our two-stage loan FDMP parameters, we set the threshold parameters $\tau$ and $\rho$ such that a good outcome, for both the applicant and lender, is when at least one loan is granted and repaid.
Table \ref{tab:appendix_two_stage_loan_marginal_welfare_pi_possible} shows that there exists a policy $\pi^\text{Fair}$ that will, in expectation, yield such welfare and cost values for every applicant.
Therefore, all applicants are considered qualified (i.e. $\Gamma=1$ for all applicants).
% therefore, we need to compute the marginal probabilities for each possible welfare value for all individuals under $\pi^\text{prime}$, which we include in table \ref{tab:appendix_two_stage_loan_marginal_welfare_pi_prime}.
Using the values shown in Table \ref{tab:appendix_two_stage_loan_marginal_welfare_pi_prime}, we see that the welfare for Prime applicants is $W_{\pi^\text{Prime}} = 2.5$, which is greater than $\tau$.
On the other hand, the welfare for Subprime applicants is $W_{\pi^\text{Prime}} = 0$ which is less than $\tau$.
Therefore, \texttt{EqOppCfUtil} evaluates as:
{%\small
\begin{align*}
    P(W_{\pi^{\text{Prime}}} \geq 1 \mid Z{=}0) &\stackrel{?}{=} P(W_{\pi^{\text{Prime}}} \geq 1 \mid Z{=}1) \\
    P(\text{Prime} \mid Z=0) &\stackrel{?}{=} P(\text{Prime} \mid Z=1) \\
        % \hspace{.3cm} \stackrel{?}{=} P\bigl( (.37\cdot 4) {+} (.09 \cdot 1) {+} &(.16 \cdot 1) {+} (.04 \cdot -2) {+} (.34 \cdot 0)\geq 1 \bigr) \\
    .34 &\neq .66 \; .
\end{align*}}%
% which is actually equivalent to \texttt{DemParWelf} (Equation \ref{eq:demparwelf_rl}), and therefore not satisfied.
% and is therefore not satisfied.
Under $\pi^\text{Prime}$, the probability that a minority applicant will have welfare above $\tau$ is $.34$.
This because the only way to have welfare above $\tau$ is to be Prime, and the probability of a minority applicant being Prime is $.34$.
Since a majority applicant has a higher probability of having welfare above $\tau$ ($.66$), $\pi^\text{Prime}$ does not satisfy \texttt{EqOppCfUtil}.

Relative to \texttt{EqOppMDPStatic}, \texttt{EqOppCfUtil} better aligns with fairness intuition as it deems the prime-only policy $\pi^\text{Prime}$ unfair since it results in a lower probability of at least one successful loan repayment for minority applicants than majority applicants.
More generally, \texttt{EqOppCfUtil} is a more robust interpretation of Equal Opportunity since it naturally allows qualification to be defined across multiple timesteps.

\begin{table*}[htb]
    \small
    \center
    \begin{tabular}{c|c|c|c|c|c|c|c|c}
        Applicant Type & $Z$ & $1^\text{st}$ Outcome & $2^\text{nd}$ Outcome & $\sum w$ & $-\sum R$ & Probability & $W_\pi = \mathbb{E}[\sum w]$ & $C_\pi = -\mathbb{E}[\sum R]$\\ \hline
        \multirow{4}{*}{Prime}    & \multirow{4}{*}{0} & Repaid    & Repaid    & +4 & -6 & $(.7)(.8)=.56$ & \multirow{4}{*}{2.5} & \multirow{4}{*}{-4.5} \\
                                  &                    & Repaid    & Defaulted & +1 & -3 & $(.7)(.2)=.14$ &                      &                       \\
                                  &                    & Defaulted & Repaid    & +1 & -3 & $(.3)(.8)=.24$ &                      &                       \\
                                  &                    & Defaulted & Defaulted & -2 &  0 & $(.3)(.2)=.06$ &                      &                       \\ \hdashline
        \multirow{2}{*}{Subprime} & \multirow{2}{*}{0} & Rejected  & Repaid    & +2 & -5 & $(1)(.7) =.70$ & \multirow{2}{*}{1.1} & \multirow{2}{*}{-4.1} \\
                                  &                    & Rejected  & Defaulted & -1 & -2 & $(1)(.3) =.30$ &                      &                       \\ \hline
        \multirow{4}{*}{Prime}    & \multirow{4}{*}{1} & Repaid    & Repaid    & +4 & -6 & $(.7)(.8)=.56$ & \multirow{4}{*}{2.5} & \multirow{4}{*}{-4.5} \\
                                  &                    & Repaid    & Defaulted & +1 & -3 & $(.7)(.2)=.14$ &                      &                       \\
                                  &                    & Defaulted & Repaid    & +1 & -3 & $(.3)(.8)=.24$ &                      &                       \\
                                  &                    & Defaulted & Defaulted & -2 &  0 & $(.3)(.2)=.06$ &                      &                       \\ \hdashline
        \multirow{2}{*}{Subprime} & \multirow{2}{*}{1} & Rejected  & Repaid    & +2 & -5 & $(1)(.7) =.70$ & \multirow{2}{*}{1.1} & \multirow{2}{*}{-4.1} \\
                                  &                    & Rejected  & Defaulted & -1 & -2 & $(1)(.3) =.30$ &                      &                       \\
    \end{tabular}
    \caption{
        The above calculations correspond to the outcomes produced by the policy $\pi^\text{Fair}$, which is the policy that rejects Subprime applicants in the first timestep, and grants loans otherwise.
        Since all applicants have $W_\pi \geq \tau$ and $C_\pi \leq \rho$, all applicants are qualified (i.e. $\Gamma=1$) when evaluating \textit{any} policy for \texttt{EqOppCfUtil}.
    }
    \label{tab:appendix_two_stage_loan_marginal_welfare_pi_possible}
\end{table*}

\begin{table*}[htb]
    \small
    \center
    \begin{tabular}{c|c|c|c|c|c|c|c|c|c}
        App Type & $Z$ & $1^\text{st}$ Outcome & $2^\text{nd}$ Outcome & $\sum w$ & $\Gamma$ & Probability & $W_\pi$ & $P(\text{AppType} \mid Z)$ & $P(W_{\pi} \geq \tau \mid Z, \Gamma{=}1)$ \\ \hline
        \multirow{4}{*}{Prime}    & \multirow{4}{*}{0} & Repaid    & Repaid    & +4 & $1$ & $(.7)(.8)=.56$ & \multirow{4}{*}{$2.5$} & \multirow{4}{*}{$.34$}  & \multirow{5}{*}{$.34$} \\ 
                                  &                    & Repaid    & Defaulted & +1 & $1$ & $(.7)(.2)=.14$ &                        &                         &                        \\
                                  &                    & Defaulted & Repaid    & +1 & $1$ & $(.3)(.8)=.24$ &                        &                         &                        \\
                                  &                    & Defaulted & Defaulted & -2 & $1$ & $(.3)(.2)=.06$ &                        &                         &                        \\ \cdashline{0-8}
        \multirow{1}{*}{Subprime} & \multirow{1}{*}{0} & Rejected  & Rejected  &  0 & $1$ & $(1)(1) =1$    & \multirow{1}{*}{$0.0$} & \multirow{1}{*}{$.66$}  &                        \\ \hline
        \multirow{4}{*}{Prime}    & \multirow{4}{*}{1} & Repaid    & Repaid    & +4 & $1$ & $(.7)(.8)=.56$ & \multirow{5}{*}{$2.5$} & \multirow{5}{*}{$.66$}  & \multirow{5}{*}{$.66$} \\
                                  &                    & Repaid    & Defaulted & +1 & $1$ & $(.7)(.2)=.14$ &                        &                         &                        \\
                                  &                    & Defaulted & Repaid    & +1 & $1$ & $(.3)(.8)=.24$ &                        &                         &                        \\
                                  &                    & Defaulted & Defaulted & -2 & $1$ & $(.3)(.2)=.06$ &                        &                         &                        \\ \cdashline{0-8}
        \multirow{1}{*}{Subprime} & \multirow{1}{*}{1} & Rejected  & Rejected  &  0 & $1$ & $(1)(1) =1$    & \multirow{1}{*}{$0.0$} & \multirow{1}{*}{$.34$}  &                        \\
    \end{tabular}
    \caption{
        Calculations for the probabilities of each possible outcome under policy $\pi^\text{Prime}$.
        E.g. under $\pi^\text{Prime}$, the probability that a $Z=0$ Prime applicant will repay loans in both timesteps, thus resulting in $+4$ welfare, is $.56$.
        % the welfare for a $Z=0$ Prime applicant is $2.5$;
        These calculations are used to produce the right-most column, which is used to evaluate \texttt{EqOppCfUtil}.
        % \texttt{EqOppCfUtil} is evaluated by comparing the two right-most columns, which represent the probability that the welfare for the given value of $Z$ exceeds the minimum welfare threshold $\tau=1$.
        Since $.34 \neq .66$, $\pi^\text{Prime}$ does not satisfy \texttt{EqOppCfUtil}.
    }
    \label{tab:appendix_two_stage_loan_marginal_welfare_pi_prime}
\end{table*}

\end{document}